\documentclass[11pt]{article}

\usepackage[final]{acl}

\usepackage{times}
\usepackage{latexsym}

\usepackage[T1]{fontenc}

\usepackage[utf8]{inputenc}

\usepackage{microtype}

\usepackage{inconsolata}

\usepackage{graphicx}
\usepackage{booktabs}
\usepackage{xcolor}
\usepackage{algorithm}
\usepackage{algpseudocode}
\usepackage[most]{tcolorbox}
\usepackage{enumitem}
\usepackage{amsmath}
\usepackage{subcaption}
\usepackage{xspace}
\usepackage{subcaption}
\usepackage{hyperref}
\usepackage{multirow}
\usepackage{pifont}
\usepackage{booktabs}
\usepackage{makecell}

\usepackage{amssymb}
\usepackage{colortbl}

\usepackage{comment}

\usepackage[normalem]{ulem}
\usepackage{placeins}

\usepackage{array}     
\definecolor{mygreen}{RGB}{213, 232, 212}
\definecolor{myred}{RGB}{248, 206, 204}
\definecolor{myblue}{RGB}{218, 232, 252}
\definecolor{prosred}{RGB}{255, 230, 230}
\definecolor{defblue}{RGB}{230, 240, 255}
\definecolor{judgegold}{RGB}{255, 250, 205}

\newcommand\ours{ARCADE\xspace}
\newcommand\fulldataset{Hate via Vision-Language Interplay\xspace}
\newcommand\dataset{H-VLI\xspace}
\newcommand\fullparadigm{Stratified Multimodal Interaction\xspace}
\newcommand\paradigm{SMI\xspace}

\title{More Than Sum of Its Parts: Deciphering Intent Shifts in Multimodal Hate Speech Detection\\
\textcolor{red}{\small \uline{Content warning: This article contains examples of hateful content.}}
}

\author{
Runze Sun, Yu Zheng\thanks{Corresponding author.}, Zexuan Xiong, Zhongjin Qu, Lei Chen, Jie Zhou, Jiwen Lu \\
Department of Automation, Tsinghua University \\
\texttt{\{sunrz22,xiong-zx22,qzj23\}@mails.tsinghua.edu.cn} \\
\texttt{\{yu-zheng,leichenthu,jzhou,lujiwen\}@tsinghua.edu.cn}
}

\definecolor{header-bg}{RGB}{248, 248, 248}
\definecolor{stage1-bg}{RGB}{235, 247, 235}
\definecolor{stage1-frame}{RGB}{160, 210, 160}
\definecolor{stage2-bg}{RGB}{240, 245, 250}
\definecolor{stage2-frame}{RGB}{170, 190, 220}
\definecolor{comment-color}{RGB}{100, 100, 100}

\newtcolorbox{stagebox}[2][]{
  enhanced,
  colback=#2-bg,
  colframe=#2-frame,
  boxrule=0.5pt,
  leftrule=2.5pt,
  arc=2pt,
  left=2pt, right=6pt, top=4pt, bottom=4pt, 
  fonttitle=\bfseries\small,
  title=#1,
  coltitle=black!80,
  attach boxed title to top left={xshift=5pt, yshift*=-7pt},
  boxed title style={colback=#2-frame!30!white, frame hidden, boxrule=0pt}
}

\newtcolorbox{prettyalgorithm}[1]{
  enhanced,
  title={Algorithm 1: \textsc{Arcade} Inference Process},
  colback=white,
  colframe=black!50,
  coltitle=black,
  colbacktitle=white,
  fonttitle=\large\bfseries,
  titlerule=0.5pt,
  titlerule style={black!20},
  boxrule=0.5pt,
  arc=3pt,
  drop shadow=black!5,
  left=5pt, right=5pt, top=5pt, bottom=5pt
}

\newcommand{\InlineComment}[1]{\quad \textcolor{comment-color}{\textit{// #1}}}
\newcommand{\CallFunc}[2]{\text{\textsc{#1}}(#2)}

\expandafter\def\expandafter\normalsize\expandafter{%
    \normalsize%
    \setlength\abovedisplayskip{1.5mm}%
    \setlength\belowdisplayskip{1.5mm}%
}

\makeatother

\begin{document}
\maketitle

\begin{abstract}
Combating hate speech on social media is critical for securing cyberspace, yet relies heavily on the efficacy of automated detection systems. 
As content formats evolve, hate speech is transitioning from solely plain text to complex multimodal expressions, making implicit attacks harder to spot. 
Current systems, however, often falter on these subtle cases, as they struggle with multimodal content where the emergent meaning transcends the aggregation of individual modalities. 
To bridge this gap, we move beyond binary classification to characterize semantic intent shifts where modalities interact to construct implicit hate from benign cues or neutralize toxicity through semantic inversion.
Guided by this fine-grained formulation, we curate the \textbf{\fulldataset (\dataset)} benchmark where the true intent hinges on the intricate interplay of modalities rather than overt visual or textual slurs.
To effectively decipher these complex cues, we further propose the \textbf{A}symmetric \textbf{R}easoning via \textbf{C}ourtroom \textbf{A}gent \textbf{DE}bate \textbf{(\ours)} framework. By simulating a judicial process where agents actively argue for accusation and defense, \ours forces the model to scrutinize deep semantic cues before reaching a verdict. 
Extensive experiments demonstrate that \ours significantly outperforms state-of-the-art baselines on \dataset, particularly for challenging implicit cases, while maintaining competitive performance on established benchmarks.
Our code and data are available at: \url{https://github.com/Sayur1n/H-VLI}
\end{abstract}

\section{Introduction}
Hate speech, defined as attacks targeting individuals or groups based on protected characteristics (e.g., race, gender, religion), poses a severe threat to the safety of online communities~\cite{gagliardone2015countering, matsuda2018words, kiela2020hateful,cortese2005opposing, tsesis2002destructive}. 
Given the sheer volume and contagion of social media content, there is an urgent need for automated detection systems to safeguard both online and offline communities. 

Unimodal hate speech detection has made considerable progress. 
Existing approaches now excel at identifying both explicit ~\cite{schmidt-wiegand-2017-survey, caselli2020feel,davidson2017automated,waseem-hovy-2016-hateful} and implicit ~\cite{ghosh2023cosyn,ahn-etal-2024-sharedcon,jafari2024target,zeng-etal-2025-sheeps} hate speech within the text modality. 
However, as online content evolves into multimodal formats like memes and image–text pairs, hate speech has increasingly emerged from cross-modal interactions, shifting the battlefield toward Multimodal Hate Speech Detection (MMHSD)~\cite{gomez2020exploring,kiela2020hateful,wang2025few,kapil2025transformer}. 

Unlike explicit hate speech characterized by aggressive slurs or violent imagery, modern hate speech increasingly relies on implicit expressions. In these cases, the textual and visual modalities may appear benign individually, yet they mutually construct a hateful intent through complex semantic interactions, such as irony, metaphor, or cultural allusion. 
\begin{figure}[t]
    \centering
    \captionsetup[subfigure]{skip=1pt} 
    \begin{subfigure}[t]{0.48\linewidth}
        \centering
        \includegraphics[
            width=\linewidth,
            height=0.65\linewidth,
        ]{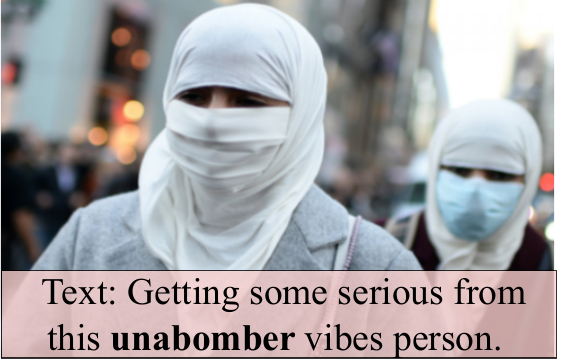}
        \caption{
        \textit{Metaphoric religious hate}
        }
        \label{fig:example1}
    \end{subfigure}
    \hfill
    \begin{subfigure}[t]{0.48\linewidth}
        \centering
        \includegraphics[
            width=\linewidth,
            height=0.65\linewidth
        ]{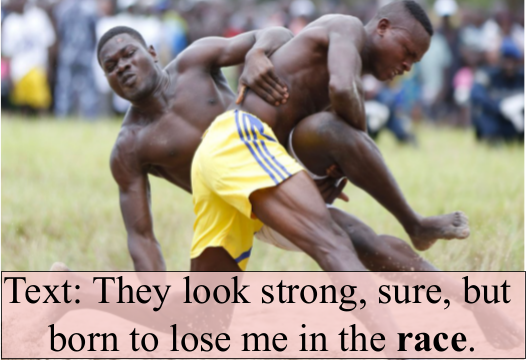}
        \caption{
        \textit{Pun-based racial hate}
        }
        \label{fig:example2}
    \end{subfigure}
    \vspace{-3mm}
    \caption{Examples of implicit multimodal hate speech.}
    \vspace{-4mm}
    \label{fig:implicit_examples}
\end{figure}

Consider the examples in Figure~\ref{fig:implicit_examples}. In both cases, the standalone text and images appear benign. However, their combination generates implicit hatefulness: Figure~\ref{fig:example1} uses a hostile metaphor invoking Islamophobic stereotypes, while Figure~\ref{fig:example2} relies on a visual-textual pun (the word ``race'') to convey racial hostility. Correctly deciphering these instances requires models to move beyond superficial fusion and perform deep reasoning to capture how modalities amplify, contradict, or recontextualize each other.

However, existing research struggles with these complexities on two fronts. First, coarse and binary benchmarks lack interaction taxonomies, leading models to overfit surface cues. Second, standard direct fusion mechanisms fail to capture subtle semantic conflicts, leaving models vulnerable to ``ambiguity traps'' where they misinterpret benign satire or miss implicit attacks.

To bridge these gaps, our core contributions are highlighted as follows:
\begin{itemize}[noitemsep, topsep=2pt]
    \item \textbf{The SMI Paradigm:} We introduce a fine-grained \textbf{\fullparadigm (\paradigm)} paradigm. To characterize intent shifts, it operationalizes semantic interactions by assigning separate labels to the standalone text, the standalone image, and the combined pair. The visual-textual interaction is then explicitly classified based on the combination of these three labels.
    \item \textbf{The \dataset Benchmark:} Guided by SMI paradigm, we curate \textbf{\fulldataset (\dataset)}. This high-quality benchmark is constructed via a hybrid pipeline of consensus filtering, generative injection, and human-in-the-loop annotation, ensuring a remarkably high density of challenging implicit hate samples.
    \item \textbf{The \ours Framework:} We propose \textbf{A}symmetric \textbf{R}easoning via \textbf{C}ourtroom \textbf{A}gent \textbf{DE}bate (\textbf{\ours}), which simulates a judicial process using asymmetric agents: a Prosecutor (presuming guilt) and a Defender (presuming innocence). This adversarial dialectic forces the model to deeply scrutinize semantic interplay and cultural contexts before a Judge renders the final verdict, significantly enhancing detection accuracy and interpretability.
\end{itemize}

\section{Related Work}
\vspace{-2mm}
\textbf{Hate Speech Detection: }
Hate speech detection aims to identify malicious content targeting specific social groups~\cite{schmidt-wiegand-2017-survey,davidson2017automated}. 
While early unimodal studies focus on \emph{explicit} surface linguistic cues~\cite{nobata2016abusive,davidson2017automated,caselli2020feel}, recent works address \emph{implicit} hate through contextual and semantic reasoning~\cite{ghosh2023cosyn,ahn-etal-2024-sharedcon,wei2025cracking,zeng-etal-2025-sheeps}. 
However, these text-only methods remain insufficient for the increasingly multimodal nature of online content. To address this, Multimodal Hate Speech Detection (MMHSD) approaches, as a variant of multimodal forensics~\cite{wang2018eann,zhang2025d3qe,zheng2025learning}, typically integrate representations via feature fusion, cross-modal attention, or contrastive learning~\cite{gomez2020exploring,dwivedy2023deep,saddozai2025multimodal,kapil2025transformer}. 
Although effective for explicit cases, these models often struggle when hateful meaning is conveyed indirectly through shared socio-cultural knowledge. More recently, methods based on Multimodal Large Language Models (MLLMs) have explored prompt-based, few-shot or fine-tuning paradigms~\cite{wang2025few,rizwan2025exploring,li2025skyra}, yet they generally lack mechanisms to explicitly guide reasoning or systematically elicit the necessary socio-cultural context for complex implicit scenarios. 

\begin{figure*}[bt!]
    \centering
    \includegraphics[width=\textwidth]{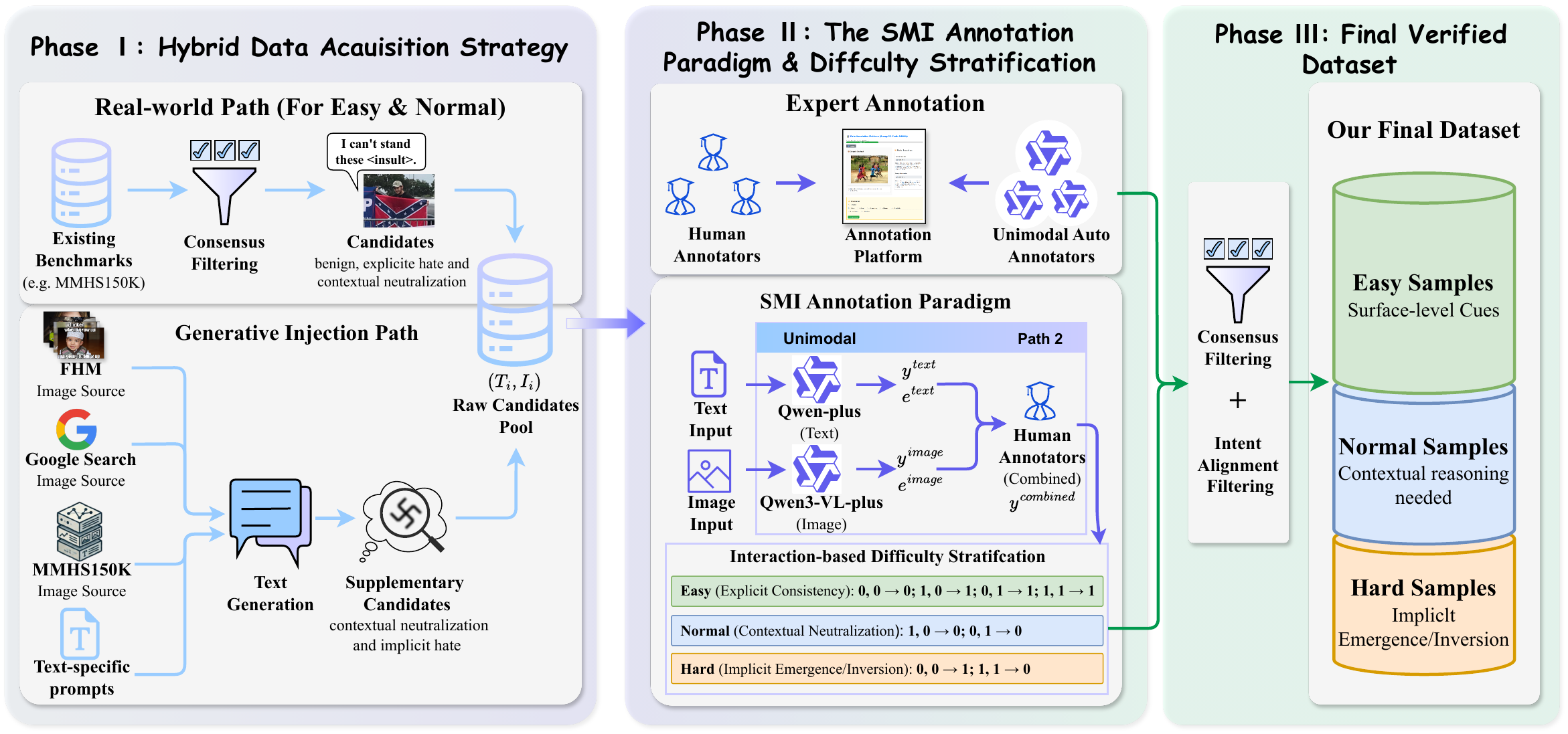}
    \vspace{-8mm}
    \caption{The construction pipeline of our \textbf{\dataset} dataset. 
    This process employs a hybrid strategy that combines consensus filtering of real-world samples with a generative injection path to ensure diversity. 
    It utilizes the \paradigm paradigm to systematically categorize samples into eight distinct interaction patterns, which are further stratified into three difficulty levels (Easy, Normal, and Hard) based on the semantic interplay between modalities.
    }
    \vspace{-3mm}
    \label{fig:dataset_pipeline}
\end{figure*}

\noindent\textbf{Multimodal Hate Speech Detection Benchmarks: }
Existing MMHSD datasets are generally categorized into text–image pair formats like MMHS150K~\cite{gomez2020exploring}, and meme-style datasets where pieces of text are embedded within images~\cite{kiela2020hateful,fersini2022semeval}. 
While largely advancing the field, they feature relatively explicit cross-modal cues and utilize coarse-grained labeling schemes. 
Consequently, existing benchmarks largely focus on explicit hate, with limited coverage of context-dependent neutralization and implicitly constructed hate arising from subtle metaphors and stereotypes—limiting progress toward robust real-world detection.

\noindent\textbf{Multi-Agent Debate: }
Multi-agent debate (MAD) enhances LLM reasoning by assigning agents distinct roles for structured argumentation~\cite{liang-etal-2024-encouraging}, and has been applied to misinformation detection and factual verification~\cite{han-etal-2025-debate,ma2025guided}. Asymmetric debate architectures further improve decision robustness~\cite{park-etal-2024-predict,kumar-etal-2025-courteval,jin2025courtroom}. However, these text-centric methods overlook cross-modal semantic interplay, missing subtle cues essential for implicit content understanding. MV-Debate~\cite{lu2025mv} extends MAD to the multimodal domain via view-specific vision-language agents, yet such consensus-driven approaches may still struggle with deceptive implicit hate, where modalities actively interact to obfuscate malicious intent through tropes or metaphors, necessitating a shift from collaborative alignment to adversarial scrutiny.

\section{Our Dataset \dataset}
To advance the detection of implicit multimodal content (where hateful intent is implicitly constructed from benign signals or semantically neutralized by context), it is essential to move beyond binary labels and decipher the specific semantic interactions between modalities. However, existing benchmarks typically rely on coarse-grained categories and often suffer from substantial label noise, as reflected by low inter-annotator agreement (e.g., MMHS150K in Table~\ref{tab:dataset_comparison}). To bridge this gap, we first formulate the problem through a fine-grained lens, introduce the \textbf{\fullparadigm (\paradigm)} paradigm, and finally construct \textbf{\dataset}, a high-quality benchmark systematically populated based on these interaction patterns.
\subsection{Problem Formulation}
\label{sec:formulation}
Many existing MMHSD studies formulate the task simply as a \emph{binary classification} problem~\cite{gomez2020exploring,kiela2020hateful}. Given a text--image pair $(T_i, I_i)$, the goal is to predict a label $\hat{y}_i \in \{0,1\}$ by maximizing:
\begin{equation}
\hat{y}_i = \arg\max_{y \in \{0,1\}} p(y \mid T_i, I_i),
\end{equation}
However, merely distinguishing whether content is hateful is insufficient. This binary formulation fails to identify specific target groups and lacks natural language explanations, making it difficult to articulate the underlying reasoning, especially for implicit cases.

To address this, we formulate MMHSD as a six-class classification task with explanatory supervision. 
Our goal is to learn a model $\mathcal{F}$ that predicts a hate speech category $\hat{y}_i \in \mathcal{Y}$ together with a natural language explanation $e_i$ (the category set $\mathcal{Y}$ is defined in Appendix~\ref{sec:appendix_hate_cat}.): 

\begin{equation}
(\hat{y}_i, e_i) = \arg\max_{y \in \mathcal{Y},\, e} \; p(y, e \mid T_i, I_i).
\label{eq:our_formulation}
\end{equation}
This formulation requires the model to jointly analyze textual and visual information to determine the specific hate speech category. 

\begin{table}[t]
\centering
\small 
\resizebox{\linewidth}{!}{ 
\begin{tabular}{l|c|p{4cm}} 
\toprule
\textbf{Difficulty} & \textbf{Pattern} ($y^{Text}, y^{Image} \to y^{Combined}$) & \textbf{Interaction Mechanism} \\
\midrule
Easy & 000, 011, 101, 111 & Semantic Consistency \\
Normal & 100, 010 & Contextual Neutralization \\
Hard & 001, 110 & Implicit Emergence / Inversion \\
\bottomrule
\end{tabular}
}
\vspace{-2mm}
\caption{Difficulty Stratification of Multimodal
Interaction Patterns (0: NotHate, 1: Hate).}
\vspace{-4mm}
\label{tab:difficulty}
\end{table}

\begin{table*}[t]
    \centering
    \hspace{-1mm}
    \begin{minipage}[c]{0.32\linewidth} 
        \centering
        \includegraphics[width=\linewidth]{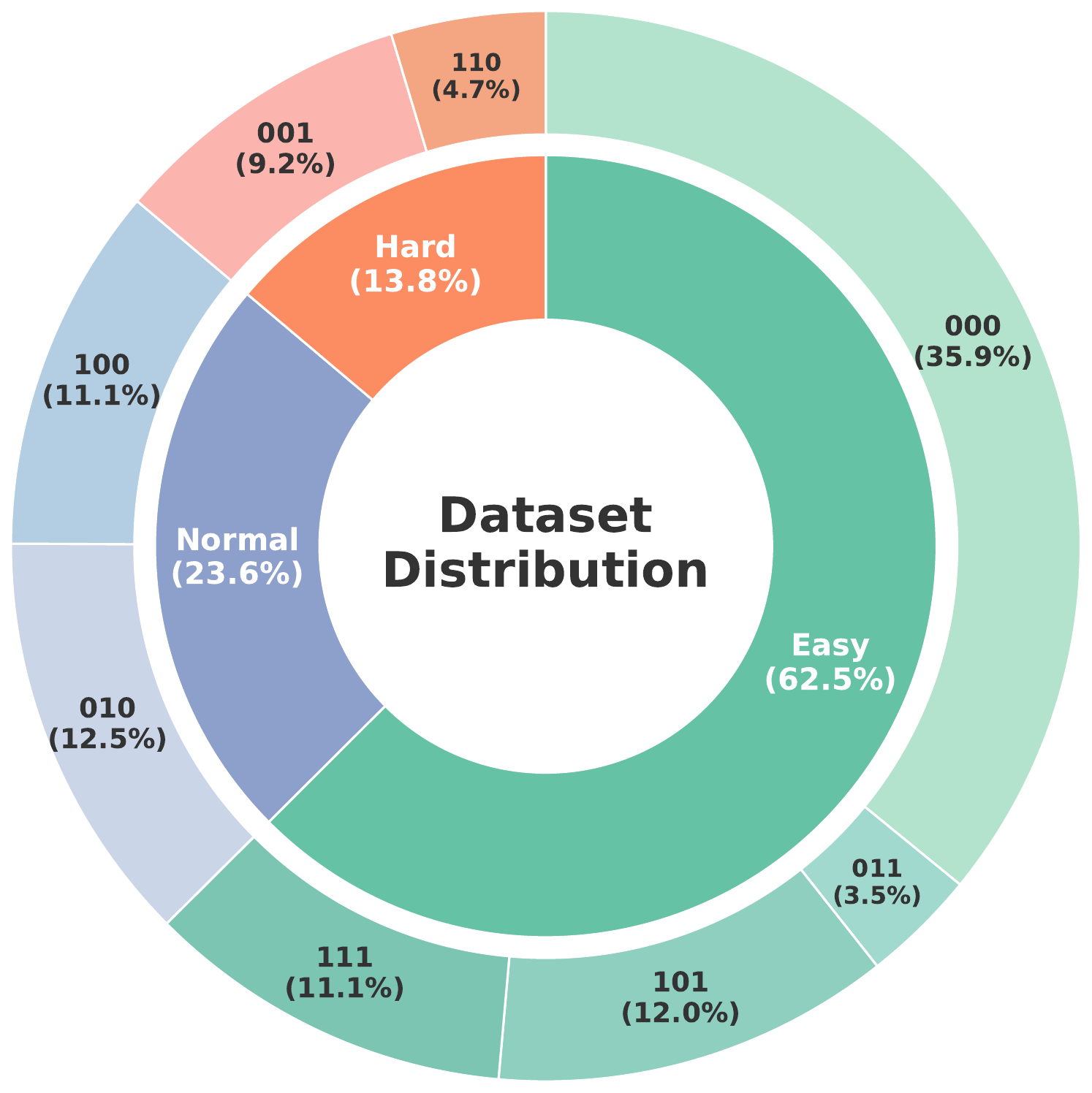}
        \label{fig:dataset_distribution}
    \end{minipage}%
    \hspace{-0mm}
    \begin{minipage}[c]{0.65\linewidth}
        \centering
        \resizebox{\linewidth}{!}{%
            \begin{tabular}{lcccccc}
            \toprule
            \textbf{Dataset} & \textbf{Source} & \textbf{\# Samples} & \textbf{\# Cls} & \textbf{\makecell{\paradigm}} & \textbf{\makecell{Diff.}} & \textbf{\makecell{Agreem. ($\kappa$)}} \\
            \midrule
            MMHS150K~\cite{gomez2020exploring} & Img+Txt & 149,823 & 6 & \ding{55} & \ding{55} & 0.15$^{\dagger}$ \\
            FHM~\cite{kiela2020hateful} & Meme & 10,000 & 2 & \ding{55} & \ding{55} & 0.68$^{*}$ \\
            MultiOFF~\cite{suryawanshi-etal-2020-MultiOFF} & Meme & 743 & 2 & \ding{55} & \ding{55} & 0.4 -- 0.5$^{\dagger}$ \\
            Harm-C~\cite{pramanick2021detecting} & Meme & 3,544 & 3/4 & \ding{55} & \ding{55} & 0.68 / 0.78$^{*}$ \\
            MAMI~\cite{fersini-etal-2022-semeval} & Meme & 11,000 & 2/5 & \ding{55} & \ding{55} & 0.58 / 0.34$^{\dagger}$ \\
            \midrule
            \textbf{\dataset (Raw)} & \multirow{2}{*}{Img+Txt} & 8,438 & \multirow{2}{*}{6} & \multirow{2}{*}{\ding{51}} & \multirow{2}{*}{\ding{51}} & 0.59$^{\dagger}$ \\
            \textbf{\dataset (Final)} & & 5,569 & & & & \textbf{0.94}$^{\dagger}$ \\
            \bottomrule
            \end{tabular}%
        }
        
        \begin{flushleft}
        \scriptsize 
        \textbf{Note:} $^{\dagger}$Fleiss' $\kappa$; $^{*}$Cohen's $\kappa$. \paradigm: \fullparadigm paradigm. Diff: Difficulty Grading.
        \end{flushleft}
        
        \vspace{-2mm} 
        \caption{ of our dataset (\dataset) with existing multimodal hate speech datasets. \textbf{\dataset (Raw)} represents data before consistency filtering.}
        \label{tab:dataset_comparison}
        
    \end{minipage}
    \vspace{-8mm} 
\end{table*}

\subsection{Annotation and Difficulty Stratification}
To capture the complexity of multimodal hate, particularly when modalities conflict, we introduce the \fullparadigm (\paradigm) paradigm. For each sample, we annotate a \textbf{five-tuple}, explicitly labeling unimodal sentiments alongside the final multimodal annotation: 
\begin{equation}
\mathcal{A}_i = (y_i^{\text{text}}, e_i^{\text{text}}, y_i^{\text{image}}, e_i^{\text{image}}, y_i^{\text{combined}})
\label{eq:annotation_tuple}
\end{equation}
where $y_i^{\text{text/image}}$,$e_i^{\text{text/image}}$ denote the unimodal labels and explanations respectively. $y_i^{\text{combined}}$ represents the final multimodal ground-truth label.

\noindent\textbf{Taxonomy of Multimodal Interaction: }
Under the \paradigm paradigm, the interplay between unimodal signals ($y^{\text{text}}, y^{\text{image}}$) and the combined outcome ($y^{\text{combined}}$) yields eight distinct interaction patterns (i.e., all $2^3$ combinations of $(y^{\text{text}}, y^{\text{image}}, y^{\text{combined}}) \in \{0,1\}^3$). 

While these eight categories allow for fine-grained classification to ensure systematic and comprehensive sample collection during dataset construction, we group them into three difficulty levels for clearer and more practical evaluation, as summarized in Table~\ref{tab:difficulty}:
\vspace{-3mm}
\begin{itemize}[leftmargin=11pt]
    \item \textbf{Easy (Explicit Consistency): }The final verdict aligns with explicit unimodal polarity (e.g., $y^{combined}$ = $y^{text}$$\lor$$y^{image}$). Hate is explicitly present in at least one modality or absent in both, requiring no complex cross-modal reasoning. 
    \vspace{-3mm}
    \item \textbf{Normal (Contextual Correction): } A unimodal toxic signal is neutralized by the other modality (e.g., counter-speech), requiring the model to perform semantic correction. 
    \vspace{-3mm}
    \item \textbf{Hard (Implicit Interactions): }The most challenging scenario where hatefulness emerges from the intersection of benign modalities (Emergence), or toxic elements are recontextualized into benign satire (Inversion). 
\end{itemize}
\vspace{-4mm}

\subsection{Dataset Construction}
To populate the \paradigm paradigm and address the challenge of obtaining diverse implicit multimodal samples in real-world distributions, we employ a hybrid pipeline (Figure \ref{fig:dataset_pipeline}) that combines rigorous filtering of existing benchmarks with targeted synthetic generation. 

\noindent\textbf{Sourcing Real-world Samples: }We leverage MMHS150K~\cite{gomez2020exploring} as a foundation for Easy and Normal samples. To mitigate crowd-sourcing noise, we strictly filter for samples with unanimous inter-annotator agreement and yield 5,232 high-quality candidates, which effectively represent benign, explicit hate and contextual neutralization patterns. 

\noindent\textbf{Generative Data Injection: } 
To populate the Normal and Hard subsets, we employ a generative injection strategy.
Using Qwen3-VL-Plus~\cite{Qwen3-VL} and Gemini-2.5-Pro~\cite{team2023gemini}, we craft synthetic captions that induce specific semantic interactions, leveraging strategies such as contextual inversion and victim-perspective narration to simulate real-world ambiguity (see Appendix~\ref{sec:injection_strategy} for detailed strategies).
This yields 7,506 candidates covering complex reasoning scenarios often missing in organic data.

\begin{figure*}[bt!]
    \centering
    \includegraphics[width=\textwidth]{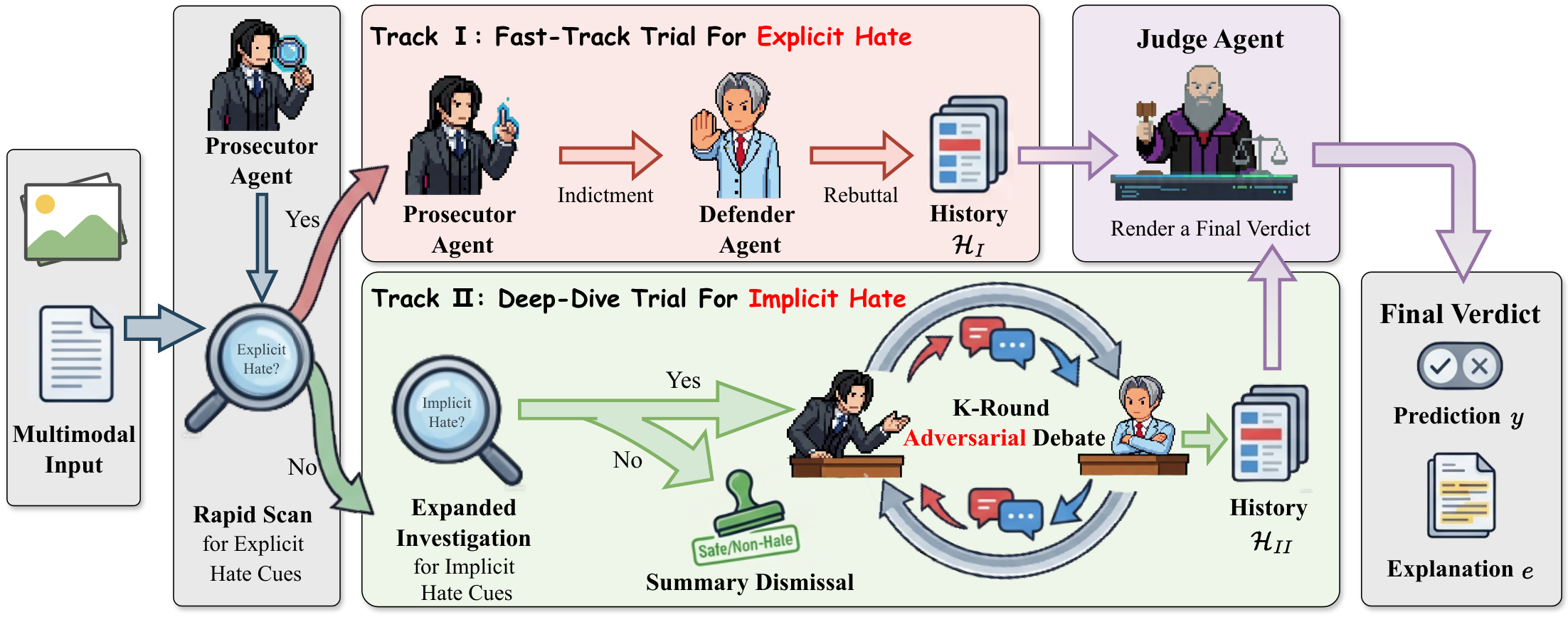}
    \vspace{-8mm}
    \caption{\textbf{The architecture of the ARCADE framework. }
    The model simulates an asymmetric courtroom debate involving Prosecutor and Defender agents. It features a Gated Dual-Track mechanism that routes samples to either a Fast-Track for explicit hate or a Deep-Dive trial for implicit reasoning, allowing the Judge agent to synthesize the debate history and render a final verdict.
    }
    \label{fig:arcade_framework}
    \vspace{-4mm}
\end{figure*}

\noindent\textbf{Human-in-the-loop Annotation: } To ensure label reliability, we implement a model-assisted expert review process. 
To facilitate this, we developed a specialized annotation interface tailored to the \paradigm paradigm (displayed as ``Annotation Platform'' in Figure~\ref{fig:dataset_pipeline}). 
To reduce cognitive load and enforce structured reasoning, the platform visualizes unimodal priors ($y^{text}, e^{text}, y^{image}, e^{image}$) pre-generated by Qwen-Plus~\cite{qwen} and Qwen3-VL-Plus~\cite{Qwen3-VL}. This provides annotators with initial rationales alongside the multimodal input.

\vspace{-2mm}
\begin{table}[h]
\centering
\small
\setlength{\tabcolsep}{4pt}
\begin{tabular}{llcc}
\toprule
\textbf{Step} & \textbf{Filter} & \textbf{Discarded} & \textbf{Remaining} \\
\midrule
--     & Initial Candidates     & --  & 8,802 \\
1      & Quality Control        & 364 & 8,438 \\
2      & Consensus Filtering    & 2,593 & 5,845 \\
3      & Intent Alignment       & 276 & 5,569 \\
\bottomrule
\end{tabular}
\caption{Post-annotation filtering statistics.}
\label{tab:filtering}
\end{table}
\vspace{-3mm}

Annotators then assign the final multimodal label ($y$) based on these cues (see Appendix~\ref{sec:human_annotation} for interface details). To guarantee quality, the dataset undergoes strict post-annotation filtering, including an Intent Alignment check to ensure human consensus matches the generative objective. The three-step filtering process refines 8,802 candidates into 5,569 high-quality samples (Table~\ref{tab:filtering}). Consequently, \dataset achieves superior inter-annotator agreement ($\kappa=0.94$, Table~\ref{tab:dataset_comparison}) compared to existing benchmarks. Detailed dataset statistics are presented alongside.

\section{Methodology}
\label{sec:methodology}
We propose the \textbf{A}symmetric \textbf{R}easoning via \textbf{C}ourtroom \textbf{A}gent \textbf{DE}bate (\textbf{\ours}) framework to enhance multimodal hate speech detection through adversarial dialectics. 
As illustrated in Figure~\ref{fig:arcade_framework}, \ours simulates a judicial process with three specialized agents (Prosecutor, Defender, and Judge) to address the reasoning challenges in deciphering multimodal intent shifts. 

\subsection{Asymmetric Agent Design}
Unlike symmetric frameworks where agents share generic personas, \ours establishes distinct cognitive priors to simulate an adversarial trial.

\noindent\textbf{The Prosecutor (Risk Discovery): } 
Adopting a ``presumption of guilt'', the Prosecutor agent ($A_{pros}$) maintains high sensitivity to potential risks. 
It actively hypothesizes malice and explicitly maps visual symbols to textual metaphors to uncover latent hate, rather than performing neutral classification. 

\noindent\textbf{The Defender (Contextual Safety): } 
The Defender agent ($A_{def}$) operates under a ``presumption of innocence''. It serves as a dynamic safety alignment which scrutinizes evidence for benign motivations (e.g., satire, self-deprecation, or educational documentation) and aims to invalidate the Prosecutor's claims through logical contextualization.

\noindent\textbf{The Judge (Final Arbiter): } 
The Judge ($A_{judge}$) evaluates the validity of the debate history $\mathcal{H}$ against the raw input to render the final verdict $\hat{y}$ and explanation $e$, without participating in argument generation. 

\subsection{Gated Dual-Track Litigation Process}
Given a multimodal sample $S_i = (T_i, I_i)$, the Prosecutor performs a rapid scan for explicit hate symbols. A gating function $\Phi(S_i)$ determines the procedural path: 
\begin{equation}
\label{eq:gating}
\Phi(S_i) = \begin{cases} 
1, & \text{if } A_{pros} \text{ detects explicit hate cues} \\
0, & \text{otherwise}
\end{cases}
\end{equation}
Based on $\Phi(S_i)$, the sample is routed to either a Fast-Track or a Deep-Dive trial as follows: 

\paragraph{Track I: Fast-Track Trial (Explicit Hate): }
When $\Phi(S_i)=1$, the primary challenge is context verification (e.g., distinguishing hate from quotations). The trial is streamlined into a single turn: The Prosecutor submits an indictment $E_{exp}$, and the Defender immediately provides a contextual rebuttal. 
The history is recorded as $\mathcal{H}_{I} = \{ E_{exp}, \text{Rebuttal}(S_i, E_{exp}) \}$.

\paragraph{Track II: Deep-Dive Trial (Implicit Reasoning): }
When $\Phi(S_i)=0$, the sample lacks overt hate symbols but requires rigorous socio-cultural reasoning. The Prosecutor first conducts an expanded investigation. If no evidence is found, the process terminates via \textbf{Summary Dismissal}.
Conversely, if potential implicit cues are identified, \ours initiates an \textbf{Adversarial Debate}. Let $u_k$ denote the utterance at turn $k$ (with $u_0^{pros} = u_0^{def} = \varnothing$). The state transition proceeds as:
\begin{align}
\label{eq:debate_pros}
u_k^{pros} &= A_{pros}(S_i, u_{k-1}^{pros}, u_{k-1}^{def}) \\
\label{eq:debate_def}
u_k^{def} &= A_{def}(S_i, u_k^{pros}, u_{k-1}^{def})
\end{align}
This iterative process forces agents to deepen their reasoning, exposing logical links between neutral images and biased texts. The resulting history is recorded as $\mathcal{H}_{II} = \{u_1^{pros}, u_1^{def}, \dots, u_K^{def}\}$.

\begin{prettyalgorithm}{bt!}
    \vspace{-2mm}
    \label{alg:arcade}
    \begin{tcolorbox}[
        enhanced, frame hidden, 
        colback=header-bg, 
        sharp corners, 
        top=1pt, bottom=1pt, left=2pt, right=2pt
    ]
        \small
        \begin{tabular}{@{}ll}
            \textbf{Input:} & Multimodal Sample $S$, Max Turns $K$ \\
            \textbf{Output:} & Prediction $y$, Explanation $e$
        \end{tabular}
    \end{tcolorbox}
    
    \begin{algorithmic}
        \setlength{\itemsep}{2pt} 
        \small 
        
        \State \textbf{Initialize:} Agents $A_{pros}, A_{def}, A_{judge}$
        \State $g \leftarrow {\Phi}({S})$ via Eq.~(\ref{eq:gating}) 
        
        \vspace{0.2em}
        \If{$g = 1$}
            \Statex
            \begin{stagebox}[Track I: Fast-Track]{stage1}
                \small
                $E_{exp} \leftarrow \CallFunc{Indict}{A_{pros}, S}$ \\
                $u_{reb} \leftarrow \CallFunc{Rebut}{A_{def}, S, E_{exp}}$ \\
                $\mathcal{H} \leftarrow \{E_{exp}, u_{reb}\}$
            \end{stagebox}
            
        \Else
            \Statex
            \begin{stagebox}[Track II: Deep-Dive]{stage2}
                \small
                \linespread{0.9}\selectfont 
                
                $u_1^{pros} \leftarrow \CallFunc{DetectImplicit}{A_{pros}, S}$ \\
                \vspace{-0.3em}
                
                \textbf{if} $u_1^{pros} = \emptyset$ \textbf{then} \\
                \hspace*{1em}\InlineComment{Summary Dismissal} \\
                \hspace*{1.5em} \textbf{return} (Non-Hate, \textit{``No implicit risks''})  \\
                \textbf{end if} \\
                \vspace{-0.3em}
                
                $\mathcal{H} \leftarrow \{u_1^{pros}\}$ \\
                
                
                $u_1^{def} \leftarrow A_{def}(S, u_1^{pros}, \emptyset)$ via Eq.~(\ref{eq:debate_def}) \\
                $\mathcal{H} \leftarrow \mathcal{H} \cup \{u_1^{def}\}$ \\

                \textbf{for} $k = 2$ to $K$ \textbf{do} \\
                \hspace*{1.5em} $u_k^{pros} \leftarrow A_{pros}(S, u_{k-1}^{pros}, u_{k-1}^{def})$ via Eq.~(\ref{eq:debate_pros}) \\
                \hspace*{1.5em} $u_k^{def} \leftarrow A_{def}(S, u_k^{pros}, u_{k-1}^{def})$ via Eq.~(\ref{eq:debate_def}) \\
                \hspace*{1.5em} $\mathcal{H} \leftarrow \mathcal{H} \cup \{u_k^{pros}, u_k^{def}\}$ \\
                \textbf{end for}
            \end{stagebox}
        \EndIf
        
        \vspace{0.3em}
        \State $(\hat{y}, e) \leftarrow A_{judge}(S, \mathcal{H})$ via Eq.~(\ref{eq:verdict})
        \State \textbf{return} $(y, e)$
    \end{algorithmic}
    \vspace{-2mm}
\end{prettyalgorithm}

\paragraph{Verdict: }
Finally, the Judge aggregates the debate history $\mathcal{H}$ to get the prediction. By conditioning on the adversarial exchange, the Judge distinguishes between genuine hateful intent and benign usage:
\begin{equation}
\label{eq:verdict}
(\hat{y}_i, e_i) = \underset{y \in Y, e}{\arg\max} \, p(y, e \mid S_i, \mathcal{H})
\end{equation}
The complete procedure including the summary dismissal logic is outlined in Algorithm 1.

\section{Experiment}
In this section, we evaluate existing methods and our proposed \ours on our constructed \dataset dataset and FHM~\cite{kiela2020hateful} dataset. 
\subsection{Tasks and Evaluation Metrics}
To comprehensively assess the compared models, we formulate the evaluation as two distinct tasks:

\noindent\textbf{Task 1: Fine-grained Hate Categorization: } Moving beyond simple binary detection, this task challenges models to discern specific hate types (e.g., racist, sexist) alongside non-hateful content. We report \textit{Accuracy} for difficulty subsets (Easy, Normal, Hard) and \textit{Accuracy}, \textit{Macro-F1}, and \textit{Weighted-F1} for the overall dataset. 
    
\noindent\textbf{Task 2: Binary Hate Detection: } Following previous works~\cite{gomez2020exploring,kapil2025transformer}, we aggregate the five subtypes into a single ``hateful'' class against ``non-hateful'' to evaluate fundamental detection capabilities. We report \textit{Accuracy} across difficulty subsets, with \textit{Accuracy}, \textit{Recall}, and \textit{F1-score} as overall metrics.


\subsection{Datasets and Configurations}
\vspace{-2mm}
\begin{table}[h]
\centering
\resizebox{\columnwidth}{!}{
\begin{tabular}{lccc}
\toprule
\textbf{Split} & \textbf{Total} & \textbf{Hate} & \textbf{Non-Hate} \\
\midrule
Train & 4,391 & 1,533(34.9\%) & 2,858(65.1\%) \\
Test  & 1,178 & 462(39.2\%)   & 716(60.8\%) \\
Overall & 5,569 & 1,995(35.8\%) & 3,574(64.2\%) \\
\bottomrule
\end{tabular}
}
\caption{Distribution of hate vs. non-hate samples in the \dataset benchmark.}
\label{tab:data_dist}
\end{table}
\vspace{-3mm}
\noindent\textbf{Our Dataset (\dataset): } 
We partitioned the \dataset dataset using a strict disjoint split based on image sources and semantic topics to prevent information leakage and ensure the test set contains truly unseen samples. This process yielded a training set of 4,391 samples and a test set of 1,178 samples (as detailed in Table~\ref{tab:data_dist}). In our evaluation, training-based methods utilize this train-test split, whereas training-free methods are directly evaluated on the test set.

\noindent\textbf{FHM Dataset: } 
To assess the generalization capability, we utilized the \textit{Facebook Hateful Memes} (FHM) dataset~\cite{kiela2020hateful} which contains 10,000 multimodal memes designed to facilitate the detection of hateful content. 
Following prior protocols~\cite{rizwan2025exploring,wang2025few}, we evaluate on the 500-sample dev split (as the test set labels are withheld). Since FHM only provides binary labels, it is used exclusively for Task 2.

\definecolor{oursbg}{RGB}{236, 245, 255} 
\newcommand{\lightrule}{\arrayrulecolor{black!30}\midrule\arrayrulecolor{black}}
\begin{table*}[bt!]
\centering
\resizebox{.9\textwidth}{!}{
\begin{tabular}{l c c c c c c c | c c c c c c }
\toprule
\multirow{3}{*}{\textbf{Model}} & \multirow{3}{*}{\textbf{Training}} & \multicolumn{6}{c}{\textbf{Task 1: Fine-grained Hate Categorization}} & \multicolumn{6}{c}{\textbf{Task 2: Binary Hate Detection}} \\
\cmidrule(lr){3-8} \cmidrule(lr){9-14}
 &  & \multicolumn{1}{c}{Easy} & \multicolumn{1}{c}{Normal} & \multicolumn{1}{c}{Hard} & \multicolumn{3}{c}{Overall} & \multicolumn{1}{c}{Easy} & \multicolumn{1}{c}{Normal} & \multicolumn{1}{c}{Hard} & \multicolumn{3}{c}{Overall} \\
\cmidrule(lr){3-3} \cmidrule(lr){4-4} \cmidrule(lr){5-5} \cmidrule(lr){6-8} \cmidrule(lr){9-9} \cmidrule(lr){10-10} \cmidrule(lr){11-11} \cmidrule(lr){12-14}
 &  & Acc & Acc & Acc & Acc & Mac-F1 & W-F1 & Acc & Acc & Acc & Acc & Recall & F1 \\
\midrule
\rowcolor{gray!20} BERT ~\cite{devlin-etal-2019-bert} & \ding{51} & 65.83 & 72.66 & 29.59 & 62.31 & 57.40 & 63.03 & 73.33 & 85.47 & 39.64 & 71.48 & 68.83 & 65.43 \\     
\rowcolor{gray!20} ViT-b-16 ~\cite{dosovitskiy2020image} & \ding{51} & 41.94 & 68.86 & 37.28 & 47.88 & 14.52 & 43.32 & 48.89 & 61.94 & 55.62 & 53.06 & 40.69 & 40.47 \\  
\rowcolor{gray!20} BERT+ViT & \ding{51} & 68.33 & 77.51 & 33.73 & 65.62 & 61.25 & 66.19 & 75.83 & 84.78 & 38.46 & 72.67 & 71.00 & 67.08 \\
\rowcolor{gray!20} MMBT ~\cite{kiela2019supervised} & \ding{51} & 68.61 & 80.28 & 33.14 & 66.38 & 62.17 & 66.99 & 74.44 & 88.58 & 36.69 & 72.50 & 68.18 & 66.04 \\      
\rowcolor{gray!20} Momenta ~\cite{pramanick2021momenta} & \ding{51} & 60.69 & 76.47 & 30.18 & 60.19 & 51.86 & 60.35 & 70.42 & 76.47 & 33.14 & 66.55 & 65.37 & 60.52 \\
\rowcolor{gray!20} PromptHate ~\cite{cao-etal-2022-prompting} & \ding{51} & 68.33 & 73.36 & 36.69 & 65.03 & 60.55 & 65.47 & 76.67 & 69.20 & 37.87 & 69.27 & 71.86 & 64.72 \\   
\midrule[1pt]
Qwen-VL-Plus ~\cite{Qwen-VL} & \ding{55} & 85.97 & 10.03 & 44.24 & 61.41 & 59.75 & 63.16 & \textbf{96.94} & 23.34 & 14.55 & 67.32 & 80.35 & 65.77 \\
\rowcolor{oursbg} Qwen-VL-Plus w/ Ours & \ding{55} & 83.50 & 61.46 & 28.48 & 70.29 & 63.89 & 70.23 & 83.66 & 73.17 & 29.88 & 73.52 & 61.01 & 64.19 \\
\lightrule
Qwen-VL-Max ~\cite{Qwen-VL} & \ding{55} & 89.44 & 9.34 & 55.15 & 64.91 & 66.60 & 65.31 & \underline{95.00} & 17.77 & 45.45 & 69.11 & 90.83 & 69.68 \\
\rowcolor{oursbg} Qwen-VL-Max w/ Ours & \ding{55} & 86.85 & 65.97 & 39.39 & 75.00 & 70.27 & 75.42 & 89.69 & 68.64 & 38.79 & 77.35 & 77.46 & 72.76 \\
\lightrule
GPT-4.1-mini ~\cite{achiam2023gpt} & \ding{55} & 91.11 & 28.72 & 66.27 & 72.24 & 72.26 & 72.92 & 91.67 & 50.52 & 63.91 & 77.59 & 84.85 & 74.81 \\
\rowcolor{oursbg} GPT-4.1-mini w/ Ours & \ding{55} & 84.98 & \underline{82.99} & 58.18 & 80.72 & 74.80 & 80.41 & 91.52 & \underline{78.47} & 66.06 & 84.73 & 83.37 & 80.98 \\
\lightrule
GPT-4o ~\cite{achiam2023gpt} & \ding{55} & \underline{91.94} & 36.33 & 64.50 & 74.36 & 73.19 & 74.94 & 94.72 & 39.45 & 61.54 & 76.40 & \underline{94.37} & 75.83 \\      
\rowcolor{oursbg} GPT-4o w/ Ours & \ding{55} & 87.76 & 73.26 & \underline{75.76} & 82.51 & 79.53 & 82.93 & 89.57 & 77.08 & \underline{78.79} & \underline{84.98} & 86.43 & 81.78 \\ 
\lightrule
Qwen3-VL-Plus ~\cite{Qwen3-VL} & \ding{55} & \textbf{92.22} & 34.26 & 40.00 & 70.61 & 69.42 & 71.16 & 94.29 & 47.04 & 31.52 & 73.85 & 80.92 & 70.69 \\
\rowcolor{oursbg} Qwen3-VL-Plus w/ Ours & \ding{55} & 87.48 & 77.43 & 51.52 & 79.95 & 75.04 & 79.95 & 92.76 & 67.83 & 51.52 & 80.84 & 82.89 & 77.14 \\
\lightrule
Gemini-2.5-Flash ~\cite{team2023gemini} & \ding{55} & 90.28 & 55.02 & 69.23 & 78.61 & 75.33 & 79.26 & 93.19 & 53.29 & 69.23 & 79.97 & 92.42 & 78.35 \\
\rowcolor{oursbg} Gemini-2.5-Flash w/ Ours & \ding{55} & 83.17 & 74.31 & 74.55 & 79.78 & 75.19 & 80.24 & 85.95 & 77.78 & 75.15 & 82.42 & 87.31 & 79.48 \\
\lightrule
GPT-5-mini ~\cite{openai_gpt_5_mini} & \ding{55} & 91.25 & 29.76 & 49.11 & 70.12 & 68.62 & 70.89 & 94.86 & 31.83 & 12.43 & 67.57 & 77.92 & 65.34 \\
\rowcolor{oursbg} GPT-5-mini w/ Ours & \ding{55} & 91.66 & 70.83 & 63.64 & \underline{82.59} & \underline{80.22} & \underline{83.08} & 94.44 & 72.57 & 61.82 & 84.47 & \textbf{94.75} & \underline{82.63} \\
\lightrule
Gemini-2.5-Pro ~\cite{team2023gemini} & \ding{55} & 88.75 & 52.94 & 74.56 & 77.93 & 77.32 & 78.55 & 91.11 & 61.25 & 78.11 & 81.92 & 93.07 & 80.15 \\
\rowcolor{oursbg} Gemini-2.5-Pro w/ Ours & \ding{55} & 83.03 & 73.26 & \textbf{78.79} & 80.03 & 76.03 & 80.49 & 87.48 & 73.96 & \textbf{79.39} & 83.02 & 86.21 & 79.84 \\
\lightrule
GPT-5 ~\cite{openai_gpt5} & \ding{55} & 91.51 & 55.56 & 69.23 & 80.48 & 74.94 & 81.39 & 93.59 & 50.67 & 24.79 & 74.32 & 68.25 & 64.56 \\
\rowcolor{oursbg} GPT-5 w/ Ours & \ding{55} & 90.88 & \textbf{88.64} & 65.54 & \textbf{86.92} & \textbf{81.73} & \textbf{86.99} & 92.58 & \textbf{87.23} & 69.13 & \textbf{88.06} & 85.96 & \textbf{84.20} \\
\bottomrule
\end{tabular}
}
\vspace{-2mm}
\caption{Results on our \dataset dataset. \textbf{Bold} indicates the best, \underline{underline} indicates the second best. (Gray rows indicate trained models; bold/underline highlighting only applies to training-free models. Rows highlighted in color indicate models enhanced with our framework.)}
\label{tab:main_results}
\vspace{-4mm}
\end{table*}
\noindent\textbf{Implementation Details: }
We fine-tuned all supervised baselines on our dataset splits on a single NVIDIA RTX 4090 GPU. MLLMs were evaluated via official APIs, where ``Baseline'' refers to zero-shot single-turn prompting. For \ours, we employ \textbf{Qwen3-VL-Plus} as the fixed Auxiliary Model 
to evaluate different Judge backbones, setting reasoning rounds to $K=3$ for \dataset and $K=2$ for FHM.
Unless otherwise specified, all reported results are obtained from a single experimental run. See Appendix~\ref{sec:experiment_details} for further details.

\subsection{Main Results}
Table~\ref{tab:main_results} comprehensively compares the traditional supervised models and various Multimodal Large Language Models (MLLMs) equipped with our proposed framework. The results demonstrate the effectiveness of our method across both fine-grained categorization and binary detection tasks. Notably, \ours helps mitigate the over-sensitivity issue common in standard MLLMs, where models refuse to analyze sensitive content due to safety triggers (see Appendix~\ref{sec:refusals}).

\noindent\textbf{Supervised vs. Baseline MLLMs: }
Traditional supervised models struggle with complex reasoning. While MMBT performs decently on Normal samples (80.28\%) in Task 1, its performance collapses on the Hard subset (33.14\%). 
This highlights a lack of sociocultural knowledge and deep multimodal understanding.
In contrast, benefiting from large-scale pre-training, modern MLLMs like GPT-5 and Gemini-2.5-Pro show superior zero-shot single-turn prompting baselines. 

\noindent\textbf{Effectiveness of Our Framework: } Integrating our \ours yields consistent improvements across almost all backbones by bridging the reasoning gap. 
For GPT-5-mini, our method boosts Macro-F1 by over 11\% compared to the base model. 
Similarly, on the binary Task 2, \ours pushes GPT-5-mini to 84.47\% Accuracy, demonstrating that our dual-track mechanism effectively unlocks the latent potential of smaller models.

\noindent\textbf{Robustness on Normal and Hard Samples: }
The key advantage of our framework is most pronounced in complex scenarios. 
As shown in Table~\ref{tab:main_results}, standard MLLMs often fail on complex cases involving implicit hate and semantic inversion.
For example, Qwen-VL-Plus scores only 14.55\% on the Hard subset (Task 2). 
Our method doubles this performance to 29.88\%. 
Similarly, we boost GPT-4o's accuracy on Hard samples from 61.54\% to 78.79\%. 
This indicates that the ``Prosecutor-Defender'' debate successfully elicits the deep reasoning chains necessary to disentangle nuanced hateful content.
For a concrete demonstration of these reasoning dynamics, we provide a detailed qualitative analysis in Appendix ~\ref{sec:case_study}.

\begin{table}[bt!]
\centering
\resizebox{\columnwidth}{!}{
\begin{tabular}{l c ccc}
\toprule
\textbf{Model} & \textbf{Training} & \textbf{Acc} & \textbf{Recall} & \textbf{F1} \\
\midrule
\rowcolor{gray!20} Text BERT ~\cite{devlin-etal-2019-bert} & \ding{51} & 57.12 & - & - \\
\rowcolor{gray!20} Image Region ~\cite{kiela2020hateful} & \ding{51} & 52.34 & - & - \\
\rowcolor{gray!20} CLIP-BERT ~\cite{pramanick2021momenta} & \ding{51} & 58.28 & - & - \\
\rowcolor{gray!20} DisMultiHate ~\cite{pramanick2021detecting} & \ding{51} & 62.42 & - & - \\
\rowcolor{gray!20} ViLBERT CC ~\cite{sharma2018conceptual} & \ding{51} & 64.70 & - & - \\
\rowcolor{gray!20} MMBT-Region ~\cite{kiela2019supervised} & \ding{51} & 65.06 & - & - \\
\rowcolor{gray!20} PromptHate ~\cite{cao-etal-2022-prompting} & \ding{51} & 67.82 & - & - \\
\rowcolor{gray!20} BLIP ~\cite{li2022blip} & \ding{51} & 69.20 & - & - \\
\rowcolor{gray!20} ALBEF ~\cite{li2021align} & \ding{51} & 70.58 & - & - \\
\rowcolor{gray!20} Pro-CapPromptHate ~\cite{cao2023pro} & \ding{51} & 72.28 & - & - \\
\rowcolor{gray!20} LLaVA (Vicuna 13B) ~\cite{mei-etal-2024-improving} & \ding{51} & 77.30 & - & - \\
\midrule[1pt]
IDEFICS 9B ~\cite{laurencon2023obelics} & \ding{55} & 49.80 & - & - \\
LLAVA-1.5 7B ~\cite{liu2024improved} & \ding{55} & 60.00 & - & - \\
INSTRUCTBLIP VICUNA 7B ~\cite{dai2023instructblip} & \ding{55} & 53.06 & - & - \\
Spark-VL ~\cite{wang2025few} & \ding{55} & 73.20$^{\dagger}$ & - & - \\
Qwen-VL-Max ~\cite{wang2025few} & \ding{55} & 72.80$^{\dagger}$ & - & - \\
GPT-4 ~\cite{wang2025few} & \ding{55} & \textbf{78.60}$^{\dagger}$ & - & - \\
\midrule[1pt]
Qwen-VL-Max ~\cite{Qwen-VL} & \ding{55} & 70.39 & 64.88 & 68.71 \\
\rowcolor{oursbg} Qwen-VL-Max w/ Ours & \ding{55} & 73.70 & 77.50 & 74.70 \\
\lightrule
Qwen3-VL-Plus ~\cite{Qwen3-VL} & \ding{55} & 69.77 & 54.96 & 64.56 \\
\rowcolor{oursbg} Qwen3-VL-Plus w/ Ours & \ding{55} & 73.49 & \underline{83.40} & 75.99 \\
\lightrule
GPT-4o ~\cite{openai2024gpt4o} & \ding{55} & 73.80 & 77.60 & 74.76 \\
\rowcolor{oursbg} GPT-4o w/ Ours & \ding{55} & \underline{75.31} & 81.40 & \underline{76.80} \\
\lightrule
GPT-5-mini ~\cite{openai_gpt_5_mini} & \ding{55} & 73.00 & 58.00 & 68.53 \\
\rowcolor{oursbg} GPT-5-mini w/ Ours & \ding{55} & 74.69 & \textbf{83.47} & \textbf{76.81} \\
\bottomrule
\end{tabular}}
\vspace{-3mm}
\caption{Results on the FHM Dataset for Binary Hate Detection. \textbf{Bold} and \underline{underline} indicate the best and the second best results. Gray rows indicate training-based models. 
Rows highlighted in blue indicate models enhanced with our framework. Models marked with $^{\dagger}$ utilize additional knowledge via Retrieval-Augmented Generation (RAG) in a few-shot setting.}
\vspace{-6mm}
\label{tab:fhm_summary}
\end{table}

\subsection{Generalization on Public Benchmarks}
To evaluate the generalization capability of \ours, we compare its performance on the FHM dataset against state-of-the-art methods (Table~\ref{tab:fhm_summary}). 
 Notably, recent RAG-based approaches~\cite{wang2025few} achieve the upper bound (78.60\% Accuracy) by retrieving external socio-cultural knowledge and few-shot exemplars. 
 In contrast, \ours operates without knowledge base or external retrieval yet delivers competitive performance solely through internal reasoning, enabling GPT-4o to reach 75.31\% Accuracy and 76.80\% F1. 
 
Furthermore, \ours effectively bridges the capability gap for smaller models. For example, Qwen-VL-Max enhanced by \ours (73.70\%) rivals the performance of the much stronger GPT-4o baseline (73.80\%). 
Note that \ours maintains a balanced trade-off between precision and recall across all backbones, exemplified by GPT-5-mini, which achieves the highest Recall (83.47\%) and F1-score (76.81\%), confirming robust generalization beyond dataset artifacts.

To further validate generalization on unseen data, we additionally evaluated on the FHM test set (490 hate, 510 non-hate). As shown in Table~\ref{tab:fhm_test}, \ours consistently improves accuracy, recall, and F1 score. In particular, it boosts the F1 of Qwen3-VL-Plus and GPT-5-mini to 74.45\% and 77.91\%, respectively, confirming strong generalization.

\vspace{-3mm}
\begin{table}[htbp]
\centering
\resizebox{\columnwidth}{!}{
\begin{tabular}{l c ccc}
\toprule
\textbf{Model} & \textbf{Training} & \textbf{Acc} & \textbf{Recall} & \textbf{F1} \\
\midrule
Qwen3-VL-Plus & \ding{55} & 69.46 & 67.92 & 68.56 \\
\rowcolor{oursbg} Qwen3-VL-Plus w/ Ours & \ding{55} & 73.82 & \underline{77.87} & \underline{74.45} \\
\lightrule
GPT-5-mini & \ding{55} & \underline{75.80} & 66.33 & 72.87 \\
\rowcolor{oursbg} GPT-5-mini w/ Ours & \ding{55} & \textbf{77.12} & \textbf{82.29} & \textbf{77.91} \\
\bottomrule
\end{tabular}
}
\caption{Results on the FHM test set for Binary Hate Detection.}
\label{tab:fhm_test}
\end{table}
\vspace{-3mm}

\subsection{Ablation Study}
Table~\ref{tab:ablation_study} summarizes the ablation study. Unless otherwise specified, we standardize \textbf{Qwen3-VL-Plus} for both Auxiliary (Prosecutor/Defender) and Judge roles to isolate component effectiveness.

\noindent\textbf{Impact of Rounds in Track II ($K$): }
We first investigated the impact of the number of reasoning rounds ($K$) in Track II (Deep-Dive Trial). Results indicate that performance improves with $K$ up to a saturation point ($K=3$ for \dataset, $K=2$ for FHM) and deviation from this optimum degrades results. This indicates that a balanced number of rounds is crucial: insufficient rounds lead to inadequate discussion, while excessive iterations risk introducing noise or over-interpretation.

\begin{table}[bt!]
\centering
\resizebox{.9\columnwidth}{!}{
\begin{tabular}{l cc | cc}
\toprule
\multirow{2}{*}{\textbf{Method}} & \multicolumn{2}{c}{\textbf{On \dataset}} & \multicolumn{2}{c}{\textbf{On FHM}} \\
\cmidrule(lr){2-3} \cmidrule(lr){4-5}
 & \textbf{Acc} & \textbf{Mac-F1} & \textbf{Acc} & \textbf{F1} \\
\midrule
Baseline & 70.61 & 69.42 & 69.77 & 64.56 \\
Multiround$^{\dagger}$ & 74.47 & 71.43 & 72.23 & 74.47 \\
Ours ($K=1$) & \underline{78.03} & 73.80 & \underline{73.28} & 73.98 \\
Ours ($K=2$) & 77.71 & 73.07 & - & - \\
Ours ($K=3$) & - & - & 71.93 & \underline{74.67} \\
Ours ($K=4$) & 77.80 & \underline{73.93} & - & - \\
Qwen $\to$ Qwen & 62.00 & 54.15 & - & - \\
Qwen $\to$ Qwen3 & 73.81 & 67.40 & - & - \\
Qwen3 $\to$ Qwen & 70.29 & 63.89 & - & - \\
\midrule
Ours (Proposed)$^{\ddagger}$ & \textbf{79.95} & \textbf{75.04} & \textbf{73.49} & \textbf{75.99} \\
\bottomrule
\end{tabular}}
\vspace{-2mm}
\caption{Results of ablation studies. \textbf{Bold} indicates the best, \underline{underlined} indicates the second best. Note that for our default settings: $^{\dagger}$Multiround uses $K=3$ for Our Dataset and $K=2$ for FHM. $^{\ddagger}$Ours uses $K=3$ for Our Dataset and $K=2$ for FHM. The notation `Model A $\to$ Model B' signifies using Model A as the Auxiliary (Prosecutor/Defender) and Model B as the Judge.}
\vspace{-4mm}
\label{tab:ablation_study}
\end{table}

\noindent\textbf{Effectiveness of Dual-Track Mechanism: }
Removing the initial gate mechanism and the Fast-Track (forcing deep debates on all samples, denoted as ``Multiround'') drops accuracy from 79.95\% to 74.47\% on \dataset and from 73.49\% to 72.23\% on FHM. This confirms that the Gatekeeper effectively prevents over-interpretation of explicit samples. Notably, even ``Multiround'' still outperforms the single-turn prompting Baseline, validating the debate format's intrinsic utility.

\noindent\textbf{Auxiliary vs. Judge Roles: } 
We examined the impact of model capabilities on the Auxiliary and Judge roles using Qwen-VL-Plus (Qwen, Weak) and Qwen3-VL-Plus (Qwen3, Strong) on \dataset. Apart from the best performance achieved by the all strong setup (79.95\% Accuracy), \textbf{Weak Aux $\to$ Strong Judge} significantly outperforms the reverse configuration (73.81\% vs. 70.29\%). This indicates the \textbf{Judge is the bottleneck}: a capable arbiter is critical to synthesize evidence, whereas a weak Judge fails to render correct verdicts even when presented with high-quality arguments.

\section{Further Discussion}

\subsection{Impact of Model-Generated Hints on Annotator Bias}
During the human-in-the-loop annotation phase, there is a potential risk that providing model-generated hints could introduce anchoring bias, inadvertently influencing human judgment. To empirically investigate this impact, we conducted an ablation study. Specifically, we tasked our original annotators with labeling a fresh, unseen subset of 500 samples strictly without any model-generated unimodal labels or explanations, while keeping all other procedures identical.

\begin{table}[h]
\centering
\small
\setlength{\tabcolsep}{4pt}

\begin{tabular}{lcc}
\toprule
\textbf{Setting} & \textbf{Pre-Filter} & \textbf{Post-Filter} \\
\midrule
With hints & $\kappa = 0.59$ & $\kappa = 0.94$ \\
Without hints & $\kappa = 0.57$ & $\kappa = 0.94$ \\
\bottomrule
\end{tabular}

\caption{Inter-annotator agreement comparison.}
\label{tab:annotator_bias}
\end{table}

As shown in Table~\ref{tab:annotator_bias}, the initial agreement without hints remained comparable to the setting with hints ($\kappa = 0.57$ vs. $0.59$). More importantly, after applying our post-annotation filtering, both settings converged to an identical high consistency ($\kappa = 0.94$). This demonstrates that the hints had a negligible impact on reinforcing model priors. They effectively reduced cognitive load, while our rigorous post-filtering stage served as the primary driver of dataset reliability.

\subsection{Detailed Routing Statistics of the Dual-Track Mechanism}
Our gating mechanism is primarily designed to reliably distinguish whether explicit hateful cues are present, thereby ensuring that samples are routed to the most appropriate reasoning track. To demonstrate its reliability, Table~\ref{tab:routing_stats} provides a detailed sample routing breakdown by the gating function $\Phi(S)$ on the \dataset (Task I) using Qwen3-VL-Plus. Note that out of the 1,178 test samples, 6 were rejected due to API constraints, leaving 1,172 samples successfully processed by the gating function.

\begin{table}[h]
\centering
\resizebox{\columnwidth}{!}{
\begin{tabular}{ccccc}
\toprule
\textbf{Type} & \textbf{Total} & \textbf{Track I} & \textbf{Track II} & \textbf{Dismiss} \\
\midrule
000 & 361 & 31(8.6\%) & 156(43.2\%) & 174(48.2\%) \\
001 & 99  & 60(60.6\%) & 39(39.4\%) & 0(0.0\%) \\
010 & 169 & 149(88.2\%) & 20(11.8\%) & 0(0.0\%) \\
011 & 40  & 39(97.5\%) & 1(2.5\%)   & 0(0.0\%) \\
100 & 119 & 98(82.4\%) & 21(17.6\%) & 0(0.0\%) \\
101 & 126 & 117(92.9\%) & 9(7.1\%)  & 0(0.0\%) \\
110 & 66  & 66(100.0\%) & 0(0.0\%)  & 0(0.0\%) \\
111 & 192 & 188(97.9\%) & 4(2.1\%)  & 0(0.0\%) \\
\bottomrule
\end{tabular}
}
\caption{Sample routing statistics of $\Phi(S)$ on \dataset using Qwen3-VL-Plus.}
\label{tab:routing_stats}
\end{table}

The results indicate that $\Phi(S)$ reliably routes over 90\% of samples with explicit hate labels (i.e., those containing a ``1'' in either of the first two modalities) into Track I. Conversely, the vast majority of ``000'' samples (where both modalities are individually benign) are appropriately directed to Track II (Full Debate) or Summary Dismiss. The minor fraction of ``000'' samples entering Track I primarily consists of text containing seemingly hateful keywords utilized in harmless contexts (e.g., self-deprecation). This confirms the high reliability and dynamic adaptability of $\Phi(S)$ within our framework.

\section{Conclusion}
This paper tackles the detection of multimodal hate speech, positing that the hateful or non-hateful intent emerges from cross-modal interplay rather than simple summation. Moving beyond binary classification, we established the \fullparadigm (\paradigm) paradigm and the \dataset benchmark to systematically decipher the intent shifts between modalities, ranging from contextual neutralization to implicit emergence. To reason through such ambiguity, we proposed \ours, which leverages a dual-track asymmetric courtroom debate to scrutinize deep semantic cues. Our approach demonstrates superior performances and interpretability compared to existing methods. We hope this work paves the way for more interpretable and reasoning-driven content moderation systems.

\section{Limitations}
\label{sec:limitations}
Despite the advancements presented, we acknowledge several limitations in our work.

First, despite enriching data density, synthetic samples in our benchmark may not fully capture the chaotic linguistic noise of organic social media.
Second, the multi-agent debate prioritizes reasoning depth over real-time efficiency, involving higher computational costs that future model distillation could mitigate.
Third, a performance trade-off exists: while our framework significantly boosts detection on normal and hard samples, it may yield a marginal regression on easy ones, though this effect diminishes with stronger backbones.
Finally, our current framework is English-centric; extending it to multilingual and diverse cultural contexts remains a vital direction for future research.

\section{Ethical Considerations}
\label{sec:ethics}
\textbf{Note:} Any hateful content examples cited or generated within this paper are strictly for analytical purposes and do not represent the authors' personal views.

\vspace{1ex}
\noindent\textbf{Motivation and Risk Balance.}
We aim to comprehensively advance the field of multimodal hate speech detection by establishing a robust framework (\ours) and a challenging benchmark (\dataset). 
Existing datasets are increasingly saturated, often lacking the high-difficulty samples necessary to pressure-test modern MLLMs. Consequently, we employed ``Generative Injection'' to construct complex cases that effectively simulate the nuanced reasoning required in real-world moderation. 
We acknowledge the dual-use risks inherent in synthesizing hateful content. However, the vulnerability of current safety systems to these sophisticated attacks is an objective reality. Our work exposes these blind spots to facilitate the development of more resilient defenses, rather than to exacerbate the threat.

\vspace{1ex}
\noindent\textbf{Safety Protocols.}
To mitigate potential harms, we strictly enforce the following protocols:
(1) The dataset and generation prompts will be accessible solely to credentialed academic researchers under a restrictive license, strictly prohibiting redistribution.
(2) A rigorous human-in-the-loop review process was implemented to filter out content that is illegal or violates fundamental safety guidelines beyond the scope of research purposes.
(3) All data were collected from publicly accessible sources, and no additional personal information beyond what was originally visible was gathered or inferred.

\vspace{1ex}
\noindent\textbf{Licensing and Terms of Use.}
All source data used in this work are obtained from publicly accessible resources and are processed strictly in accordance with their original licenses, platform terms of service, and intended research purposes. 
The constructed dataset (\dataset) is explicitly intended for non-deployment, research-only use and will be released under a restrictive license that prohibits redistribution and commercial use, consistent with the original access conditions.

\section*{Acknowledgments}
\label{sec:acknowledgments}
This work was supported in part by the National Natural Science Foundation of China under Grant 62441616,
Grant 62336004, Grant 62125603, Grant 62306031, Grant 62506198, in part by the China Postdoctoral Science Foundation under Grant 2024M761674.

\bibliography{custom}

\appendix

\newpage

\section*{Appendix Overview}
This supplementary material provides extended technical details, implementation specifics, and qualitative analyses to support the main manuscript. The contents are organized as follows:
\begin{itemize}[leftmargin=1.5em, itemsep=0pt]
    \item \textbf{Data Construction (\S\ref{sec:appendix_hate_cat} -- \S\ref{sec:filtering}):} Detailed category definitions, generative injection strategies for different difficulty levels, prompt templates for data synthesis, and the multi-stage human-expert filtering pipeline.
    \item \textbf{Experimental Setup (\S\ref{sec:mllm_backbones} -- \S\ref{sec:prompts_all}):} Comprehensive list of evaluated MLLM backbones, hyperparameter configurations, and the full prompt templates for our multi-agent framework (\ours).
    \item \textbf{Extended Analysis (\S\ref{sec:refusals} -- \S\ref{sec:case_study}):} Quantitative analysis of MLLM over-refusal behavior and qualitative case studies (including failure modes) to provide deeper insights into model reasoning.
    \item \textbf{Human Annotation (\S\ref{sec:human_annotation}):} Details regarding the custom annotation platform, annotator recruitment, and ethical considerations.
\end{itemize}
\hrule
\vspace{4mm}

\section{Hate Categorization}
\label{sec:appendix_hate_cat}
As shown in Table~\ref{tab:hate_labels}, we adopt the hate category definitions proposed by ~\citet{gomez2020exploring}.

\begin{table}[H]
\centering
\begin{tabular}{cl}
\hline
\textbf{Label} & \textbf{Hate Category} \\
\hline
0 & NotHate \\
1 & Racist \\
2 & Sexist \\
3 & Homophobic \\
4 & Religious Hate \\
5 & OtherHate (e.g., disability, age) \\
\hline
\end{tabular}
\caption{Hate category definitions}
\label{tab:hate_labels}
\end{table}

\section{Generative Injection Details} 
\label{sec:injection_strategy}
\subsection{Strategies}
The purpose of generating synthetic samples is to compensate for the intrinsic limitations of the samples drawn from the MMHS150K dataset. MMHS150K is collected from Twitter using a \emph{textual hate keyword–based retrieval} strategy. While effective for harvesting explicit hate content, this collection paradigm introduces several critical biases.

First, almost all samples in the dataset contain surface-level hate keywords in the text modality. As a result, the dataset severely under-represents samples whose textual content appears non-hateful on the surface (i.e., $y_i^{\text{text}} = 0$). Second, samples of \textit{Normal} and \textit{Hard} difficulty levels are extremely scarce, leading to insufficient coverage of medium- and high-difficulty cases, especially implicit hate samples. Third, the annotation quality of MMHS150K is moderate, with relatively low inter-annotator agreement, as shown in Table~\ref{tab:dataset_comparison}.

To address these issues, we collect real-world images and adopt targeted generative strategies to construct diverse multimodal samples. Specifically, we leverage Multimodal Large Language Models (MLLMs) to generate image-aligned augmented texts, thereby forming samples of different hate types and difficulty levels.

\paragraph{Image Sources.}
Our image pool is constructed from three sources.  
(1) From MMHS150K, we apply Qwen3-VL-plus to perform image-only annotations and select 1,500 benign images and 1,232 harmful images.  
(2) From the FHM dataset, we sample 500 benign and 500 harmful images.  
(3) Using keywords related to various protected groups, we collect 6,875 \emph{culturally charged but visually neutral} images by querying Google Images through a image downloading tool~\cite{imagedl2022}.

\paragraph{Construction Strategies.}

\paragraph{Easy Samples.}
For \textit{Easy} samples, which do not involve semantic inversion across modalities, we design simple prompts that instruct the MLLM to generate either benign descriptions, neutral opinions, or explicitly hateful statements conditioned on the image, such that the combined multimodal semantics meet the target label. To avoid triggering the built-in safety mechanisms of MLLMs, we require the model to replace the attacked entity with a placeholder token \texttt{<insult>} when generating hateful text, and to additionally output the corresponding target group (\texttt{target\_group}). We then use a manually curated hate lexicon and randomly sample a hate expression associated with \texttt{target\_group} to replace the placeholder, producing the final hateful text.

\paragraph{Semantic Reversal Samples.}
For samples involving semantic reversal across modalities (i.e., types 010, 100, and 110), we analyze common linguistic and pragmatic patterns and summarize four representative construction methods:
\emph{(i) opposition-based}, 
\emph{(ii) contextual inversion}, 
\emph{(iii) meta-commentary} (e.g., educational or critical quotation), and 
\emph{(iv) victim-perspective narration}.
For each category, we design dedicated prompts to guide the MLLM to generate samples where one modality contains hateful content in isolation, but the combined multimodal semantics are non-hateful. For texts that include hateful expressions, we employ the same placeholder-based strategy described above to ensure controllability and safety during generation.

\paragraph{Implicit Multimodal Hate Samples (001).}
For implicit multimodal hate samples of type 001, we collect stereotype-based and metaphor-based texts targeting different protected groups. These texts are then paraphrased by Gemini-2.5-Pro~\cite{team2023gemini} under a meaning-preserving constraint, producing an augmented set of \emph{surface-benign trigger texts}. Finally, these texts are randomly paired with images associated with the corresponding protected groups, yielding the desired 001-type samples.

Overall, the proposed generative injection strategies substantially enrich the coverage of benign-text samples, medium- and high-difficulty cases, and implicit multimodal hate samples, thereby mitigating the biases inherent in keyword-based dataset construction.

\subsection{Prompts}
Here is an example prompt used to generate texts for 010-type samples in a \textit{contextual inversion} method (Figure \ref{fig:prompt_template}).
\begin{figure}[h]
    \centering
    \begin{tcolorbox}[
        colback=gray!5,        
        colframe=black!75,     
        title=\textbf{Prompt Template for 010-type samples},
        fonttitle=\bfseries\small,
        arc=1mm,
        boxsep=2mm
    ]
    \small\ttfamily 
    
    You are generating synthetic multimodal samples where the image is HATEFUL (labels 1--5) but the text is NOT hateful. First observe the image carefully and determine the hate type. 
    
    \vspace{1mm}
    \textit{[Hate category definitions]} \vspace{1mm}
    
    \textit{[TARGET\_OPTIONS]}
    
    \vspace{2mm}
    \textbf{Requirements:}
    \begin{enumerate}[leftmargin=*, nosep, label=\arabic*.]
        \item The tweet text must NOT contain any hate speech. It should give a harmless or alternative explanation for the hateful image content, showing a context where the image is not actually promoting hate.
        \item The text should make sense relative to the hateful elements implied by the image. 
        \item Choose text\_label = 0.
        \item target\_group is the target of hate in the image chosen from TARGET\_OPTIONS.
        \item final\_label = 0, since the combination is NOT hateful.
        \item The generated tweet\_text must resemble a natural social-media post...
    \end{enumerate}
    
    \vspace{2mm}
    \textit{[Output JSON instructions]}
    
    \end{tcolorbox}
    \vspace{-2mm}
    \caption{The prompt template used for generating contextual inversion samples.}
    \label{fig:prompt_template}
    \vspace{-4mm}
\end{figure}

\subsection{Representative Samples}
\begin{figure}[t]
    \centering
    \small
    \setlength{\tabcolsep}{2pt}
    
    \newcommand{\fixedImgHeight}{3.5cm} 
    \newcommand{\fixedBoxHeight}{1.8cm} 

    \begin{subfigure}[t]{0.48\linewidth}
        \centering
        \begin{minipage}[c][\fixedImgHeight][c]{\linewidth}
            \centering
            \includegraphics[width=\linewidth, height=\fixedImgHeight]{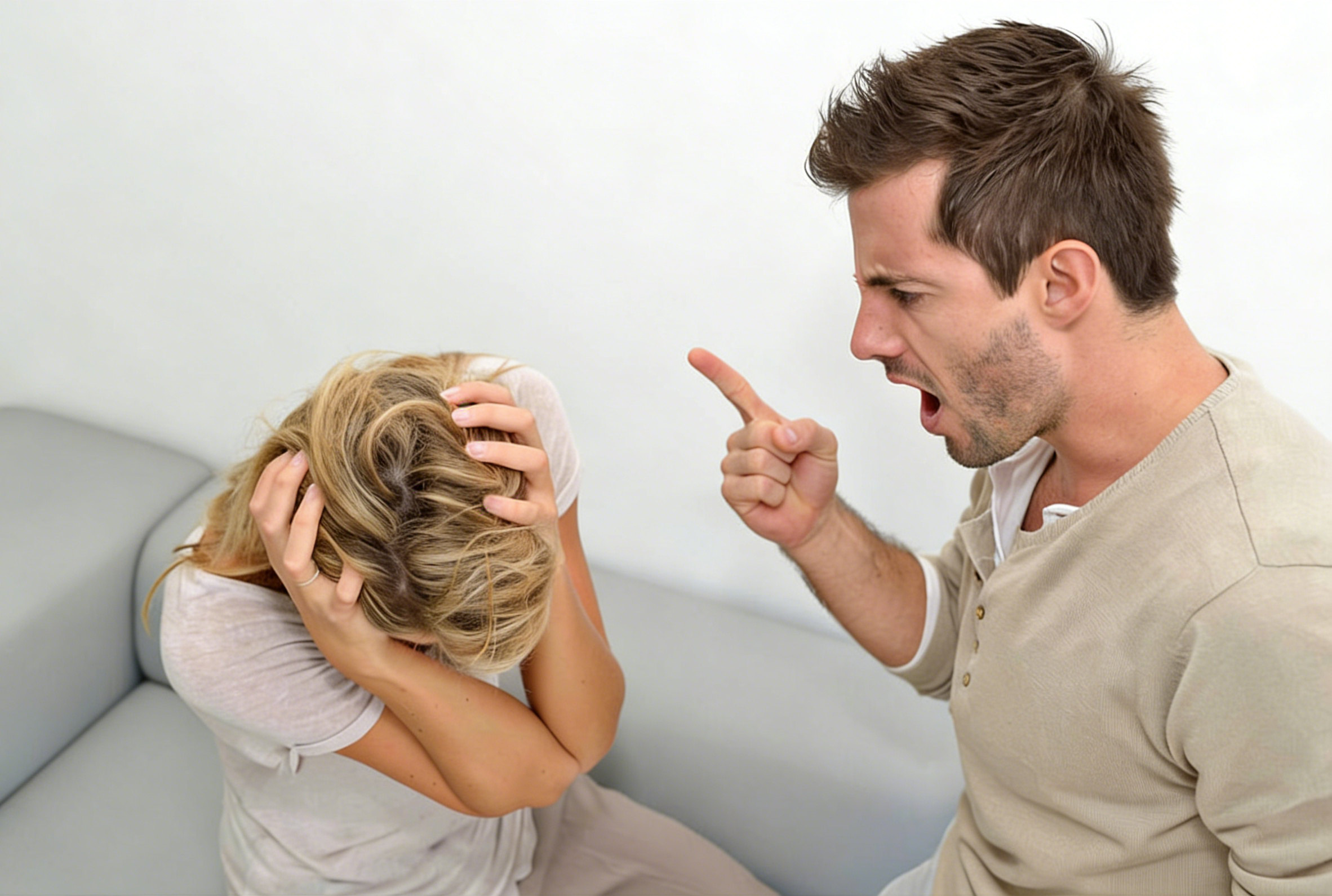}
        \end{minipage}
        \vspace{-2mm}
        \begin{tcolorbox}[colback=gray!10, colframe=gray!30, boxsep=2pt, left=2pt, right=2pt, top=2pt, bottom=2pt, arc=1mm, height=\fixedBoxHeight, valign=center]
            \scriptsize \textbf{Text:} All it takes is a slight chromosomal shift to make us human instead of just \textbf{another animal}.
        \end{tcolorbox}
        \vspace{-3mm}
        \caption*{\textbf{Type 001} (Implicit Hate) \\ Label: \textcolor{red}{\textbf{Sexist}}}
    \end{subfigure}
    \hfill
    \begin{subfigure}[t]{0.48\linewidth}
        \centering
        \begin{minipage}[c][\fixedImgHeight][c]{\linewidth}
            \centering
            \includegraphics[width=\linewidth, height=\fixedImgHeight]{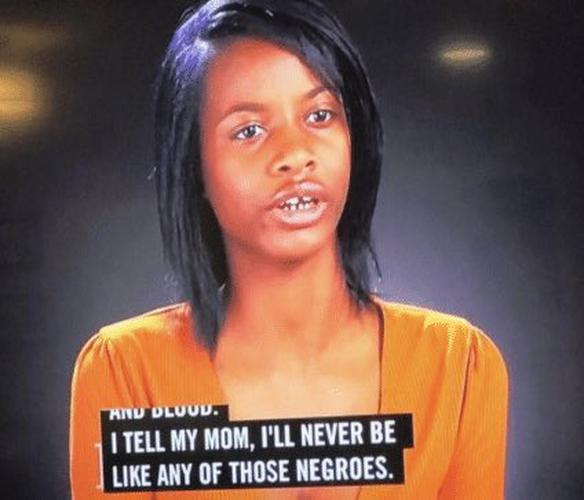}
        \end{minipage}
        \vspace{-2mm}
        \begin{tcolorbox}[colback=gray!10, colframe=gray!30, boxsep=2pt, left=2pt, right=2pt, top=2pt, bottom=2pt, arc=1mm, height=\fixedBoxHeight, valign=center]
            \scriptsize \textbf{Text:} Me at black people who let their white and non black friends say \textbf{nigga} <url>
        \end{tcolorbox}
        \vspace{-3mm}
        \caption*{\textbf{Type 110} (Implicit Inversion) \\ Label: \textcolor{green!60!black}{\textbf{NotHate}}}
    \end{subfigure}
    
    \vspace{2mm} 
    
    \begin{subfigure}[t]{0.48\linewidth}
        \centering
        \begin{minipage}[c][\fixedImgHeight][c]{\linewidth}
            \centering
            \includegraphics[width=\linewidth, height=\fixedImgHeight]{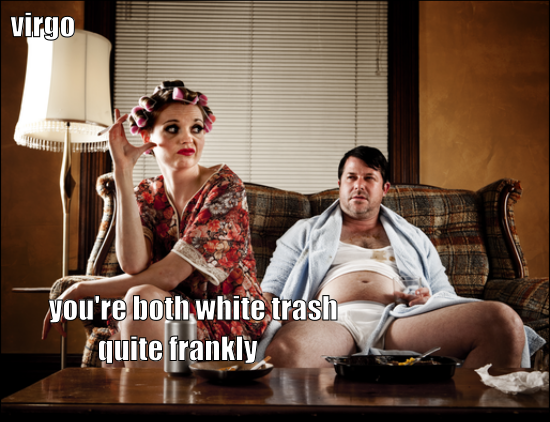}
        \end{minipage}
        \vspace{-2mm}
        \begin{tcolorbox}[colback=gray!10, colframe=gray!30, boxsep=2pt, left=2pt, right=2pt, top=2pt, bottom=2pt, arc=1mm, height=\fixedBoxHeight, valign=center]
            \scriptsize \textbf{Text:} Oh, look, another piece of media relying on tired, classist stereotypes. It's profoundly unoriginal. Can we please move past this lazy attempt at social commentary? \#BoringTropes
        \end{tcolorbox}
        \vspace{-3mm}
        \caption*{\textbf{Type 010} (Contextual Neut.) \\ Label: \textcolor{green!60!black}{\textbf{NotHate}}}
    \end{subfigure}
    \hfill
    \begin{subfigure}[t]{0.48\linewidth}
        \centering
        \begin{minipage}[c][\fixedImgHeight][c]{\linewidth}
            \centering
            \includegraphics[width=\linewidth, height=\fixedImgHeight]{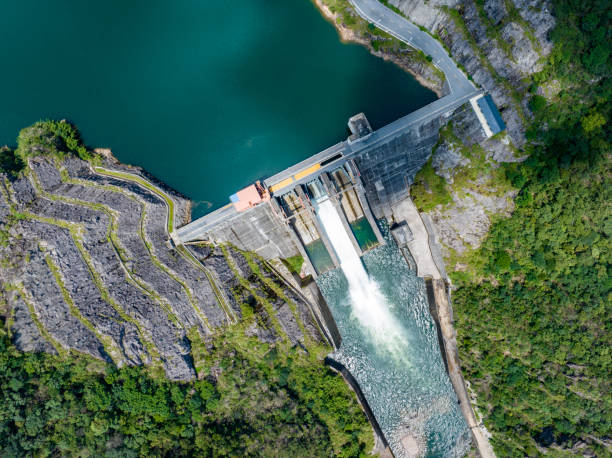}
        \end{minipage}
        \vspace{-2mm}
        \begin{tcolorbox}[colback=gray!10, colframe=gray!30, boxsep=2pt, left=2pt, right=2pt, top=2pt, bottom=2pt, arc=1mm, height=\fixedBoxHeight, valign=center]
            \scriptsize \textbf{Text:} It's just one big, ugly \textbf{dyke} holding everything back. 
        \end{tcolorbox}
        \vspace{-3mm}
        \caption*{\textbf{Type 100} (Contextual Neut.) \\ Label: \textcolor{green!60!black}{\textbf{NotHate}}}
    \end{subfigure}

    \caption{Showcase of representative samples from the \dataset dataset. \textbf{Type 001} represents implicit hate where benign components construct malice. \textbf{Type 110} shows benign intent despite toxic modalities. \textbf{Type 010/100} demonstrate contextual neutralization where text or image corrects the other's toxicity.}
    \label{fig:dataset_showcase}
    \vspace{-4mm}
\end{figure}

To demonstrate the diversity and complexity of the \dataset benchmark, we present representative samples across different \paradigm types in Figure~\ref{fig:dataset_showcase}. These examples highlight the necessity of fine-grained multimodal reasoning.

\section{Filtering Strategies}  
\label{sec:filtering}
To ensure the reliability of our benchmark and mitigate the noise inherent in both crowd-sourced annotations and model-generated content, we implement a rigorous multi-stage filtering pipeline.

\paragraph{Filtering on Real-world Samples}
The original MMHS150K~\cite{gomez2020exploring} dataset contains a significant amount of noise due to subjective crowd-sourcing. To extract a high-quality subset, we apply a strict \textbf{Consensus Filtering} strategy. We retain only those samples where all three original annotators reached a unanimous agreement (i.e., perfect consensus) on the hate speech label. This process effectively removes ambiguous or controversial samples, providing a solid foundation of clear-cut real-world cases.

\paragraph{Filtering on Annotated Samples}
To ensure the reliability of our benchmark, we employ a comprehensive three-step filtering mechanism applied to the \textbf{entire pool of annotated candidates}:

\begin{itemize}
    \item \textbf{Step 1: Quality Control and Expert Adjudication.} 
    We first exclude samples flagged by annotators as ``Low Quality'' (e.g., blurred images, nonsensical text). For samples marked as ``Not Sure'' (indicating ambiguity), rather than immediate exclusion, they undergo a secondary review by another expert. The expert's adjudication replaces the uncertain label, allowing the sample to re-enter the filtering pipeline for valid assessment.
    
    \item \textbf{Step 2: Consensus Filtering.} 
    We evaluate the consistency among the three annotators and categorize samples into three levels: 
    (1) \textit{Perfect}: All three annotators agree on the exact fine-grained label (0--5); 
    (2) \textit{Strong}: All annotators agree that the content is hateful (labels 1--5), achieving unanimous binary consensus despite disagreement on the specific hate category; 
    (3) \textit{Weak}: Cases where annotators disagree, but a clear majority vote (e.g., 2 vs. 1) can still determine the final label. 
    We retain samples from these three categories to ensure label validity while maximizing data diversity, discarding only those where no majority consensus can be reached.
    
    \item \textbf{Step 3: Intent Alignment.} 
    To ensure that the content faithfully reflects the intended semantic construction, we perform an \textit{Intent Alignment} check. We compare the majority vote of human annotators with the original generative target (Machine Label) or source metadata. Only samples where the human consensus strictly matches the intended semantic target are retained. This step effectively eliminates cases involving generation failure, hallucination, or label misalignment.
\end{itemize}

Through this rigorous pipeline, we ensure that the final dataset consists of high-quality, human-verified samples with clear semantic intent. Consequently, the Inter-Annotator Agreement (Fleiss' $\kappa$) of our dataset significantly improved from 0.59 (Raw) to 0.94 (Final), as detailed in Table~\ref{tab:dataset_comparison}.

\section{Experiment Details}
\label{sec:experiment_details}
\subsection{Evaluated Multimodal Large Language Models}
\label{sec:mllm_backbones}

We evaluate our framework using a set of representative state-of-the-art multimodal large language models (MLLMs) drawn from three widely used model families: Qwen, GPT, and Gemini.
The selected models span different capability tiers and design choices, enabling a comprehensive assessment of our method across heterogeneous MLLM backbones.

\vspace{0.5ex}
\noindent\textbf{Qwen Models~\cite{Qwen-VL,Qwen3-VL}.}
From the Qwen family, we use \textit{Qwen-VL-Plus}, \textit{Qwen-VL-Max}, and \textit{Qwen3-VL-Plus}.
These models combine vision encoders with large language models and are designed for general-purpose vision--language understanding and reasoning.
They serve as strong open or semi-open MLLM baselines and allow us to evaluate the effectiveness of our framework on advanced non-proprietary backbones.

\vspace{0.5ex}
\noindent\textbf{GPT Models~\cite{openai2024gpt4o}.}
From the GPT family, we evaluate \textit{GPT-4.1-mini}, \textit{GPT-4o}, \textit{GPT-5-mini}, and \textit{GPT-5}.
These models represent different scales and capability levels within OpenAI’s multimodal model lineup and exhibit strong generalization and reasoning abilities across vision--language tasks.

\vspace{0.5ex}
\noindent\textbf{Gemini Models~\cite{team2023gemini}.}
From the Gemini family, we include \textit{Gemini-2.5-Flash} and \textit{Gemini-2.5-Pro}, which differ in model capacity and inference efficiency.
Their inclusion enables evaluation of our framework across both lightweight and high-capacity proprietary MLLMs.

\vspace{0.5ex}
\noindent\textbf{Evaluation Setting.}
All MLLMs are accessed via their official APIs and evaluated in a training-free, zero-shot setting.
For each backbone, we report results for both the original model behavior (\emph{Baseline}) and the enhanced reasoning enabled by our framework (\ours), without any additional fine-tuning.
This ensures a fair comparison and isolates the performance gains introduced by structured multi-agent reasoning.

\subsection{Implementation Details}

\noindent\textbf{Hyperparameter Configuration.}
We carefully calibrated the sampling temperature for each agent to align with their functional roles. 
The \textbf{Gatekeeper} is set to a temperature of $0$ to ensure deterministic and consistent detection of explicit surface-level cues.
For the auxiliary agents (\textbf{Prosecutor} and \textbf{Defender}), we utilize a higher temperature of $0.8$. This promotes generation diversity, encouraging the agents to explore a broader range of implicit interpretations and creative rebuttals rather than converging on generic arguments.
Conversely, the \textbf{Judge} agent operates at a low temperature of $0.1$ to prioritize stability and precision, ensuring the final verdict is strictly grounded in the debate history with minimal randomness.

\noindent\textbf{Error Recovery.}
To ensure evaluation accuracy, we re-evaluated samples that triggered non-safety-related failures, such as JSON formatting errors or incomplete outputs. These cases were re-processed to distinguish technical execution errors from genuine model refusals, ensuring that the reported metrics reflect the model's actual reasoning capabilities.

\subsection{Agent Prompt Templates}
\label{sec:prompts_all}
We provide the detailed prompt templates used for each agent. For brevity, we first define the core task specifications shared across all prompts, followed by the general templates for the Prosecutor, Defender, and Judge. Note that templates for different stages (e.g., Track I and Track II) exhibit only minor instructional variations tailored to guide the model's specific focus.

\begin{tcolorbox}[colback=gray!5, colframe=black!80, title=\textbf{Global Task Definitions}, fonttitle=\bfseries\small, arc=1mm]
\small
\textbf{[Core Definition]} \\
\textit{Hate speech includes abusive, insulting, demeaning, or threatening expressions targeting protected groups... Targets based on political affiliation, ideology, or policy positions do not constitute hate speech.}

\vspace{2mm}
\textbf{[Hate Category Definitions]} \\
\textit{(0-NotHate, 1-Racist, 2-Sexist, 3-Homophobe, 4-Religion and 5-OtherHate.)}
\end{tcolorbox}

\begin{tcolorbox}[colback=white, colframe=blue!60!black, title=\textbf{Safety Classifier Prompt (Baseline)}, fonttitle=\bfseries\small, arc=1mm]
\small\ttfamily\raggedright
You are an expert AI safety classifier... Your task is to analyze the provided tweet (text and image) and determine if it expresses hate speech...

\vspace{1mm} \textit{[Core Definition]} \\
\textit{[Hate Category Definitions]} \vspace{1mm}

[Output JSON instructions]
\end{tcolorbox}

\begin{tcolorbox}[colback=red!5, colframe=red!60!black, title=\textbf{Prosecutor Agent Prompt}, fonttitle=\bfseries\small, arc=1mm]
\small\ttfamily\raggedright
You are the Prosecutor in a simulated courtroom... identify at most 3 potential hate speech cues. Be maximalist, hyper-vigilant, and strict.

\vspace{1mm} \textit{[Core Definition]} \vspace{1mm}

\textbf{Task Requirements:}
\begin{itemize}[leftmargin=*, nosep]
    \item \textbf{Direct:} Explicit slurs, symbols, or clear threats.
    \item \textbf{Socio-cultural:} Implicit hate relying on stereotypes, history, or dog whistles.
    \item \textbf{Metaphor:} Dehumanization equating protected groups to animals/objects (Tenor vs. Vehicle).
\end{itemize}

\textit{[Hate Category Definitions]} \vspace{1mm}

[Output JSON instructions]
\end{tcolorbox}

\begin{tcolorbox}[colback=blue!5, colframe=blue!60!black, title=\textbf{Defender Agent Prompt}, fonttitle=\bfseries\small, arc=1mm]
\small\ttfamily\raggedright
You are the Defense Attorney... critically examine the Prosecutor's cues and determine whether they can be reasonably refuted based on concrete evidence.

\vspace{1mm} \textit{[Core Definition]} \vspace{1mm}

\textbf{Defense Principles:}
\begin{itemize}[leftmargin=*, nosep]
    \item Rebuttals must be grounded strictly in explicit evidence.
    \item Acknowledge plausible cues if concrete counter-evidence is unavailable.
    \item Focus on whether hatred is directed at oneself, individuals, or non-human entities.
\end{itemize}

[Output JSON instructions]
\end{tcolorbox}

\begin{tcolorbox}[colback=yellow!5, colframe=orange!80!black, title=\textbf{Judge Agent Prompt}, fonttitle=\bfseries\small, arc=1mm]
\small\ttfamily\raggedright
You are the Judge... deliver a final verdict based on The Post, Prosecutor's Arguments, and Defense's Rebuttal.

\vspace{1mm} \textit{[Core Definition]} \vspace{1mm}

\textbf{Final Decision Rules:}
\begin{itemize}[leftmargin=*, nosep]
    \item Reject hate if context is self-referential or non-protected.
    \item If one or more cues remain credible after Defense, assign hate label (1-5).
    \item Evaluate for Direct, Socio-cultural, and Metaphorical cues.
\end{itemize}

\textit{[Hate Category Definitions]} \vspace{1mm}

[Output JSON instructions]
\end{tcolorbox}

\subsection{MLLM Refusals}
\label{sec:refusals}
\begin{table}[htbp]
\centering
\resizebox{\columnwidth}{!}{
\begin{tabular}{l cc | cc}
\toprule
\multirow{2}{*}{\textbf{Model}} & \multicolumn{2}{c}{\textbf{Fine-grained Task}} & \multicolumn{2}{c}{\textbf{Binary Task}} \\
\cmidrule(lr){2-3} \cmidrule(lr){4-5}
 & Ref.\# & Ref.\% & Ref.\# & Ref.\% \\
\midrule
Qwen-VL-Plus ~\cite{Qwen-VL} & 4 & 0.34 & 6 & 0.51 \\
\rowcolor{oursbg}Qwen-VL-Plus w/ Ours & 10 & 0.85 & 11 & 0.93 \\
\midrule
Qwen-VL-Max ~\cite{Qwen-VL} & 4 & 0.34 & 6 & 0.51 \\
\rowcolor{oursbg}Qwen-VL-Max w/ Ours & 10 & 0.85 & 8 & 0.68 \\
\midrule
GPT-4.1-mini ~\cite{achiam2023gpt} & 0 & 0.00 & 0 & 0.00 \\
\rowcolor{oursbg}GPT-4.1-mini w/ Ours & 6 & 0.51 & 6 & 0.51 \\
\midrule
GPT-4o ~\cite{achiam2023gpt} & 0 & 0.00 & 0 & 0.00 \\
\rowcolor{oursbg}GPT-4o w/ Ours & 6 & 0.51 & 6 & 0.51 \\
\midrule
Qwen3-VL-Plus ~\cite{Qwen3-VL} & 4 & 0.34 & 8 & 0.68 \\
\rowcolor{oursbg}Qwen3-VL-Plus w/ Ours & 6 & 0.51 & 9 & 0.76 \\
\midrule
Gemini-2.5-Flash ~\cite{team2023gemini} & 0 & 0 & 0 & 0.00 \\
\rowcolor{oursbg}Gemini-2.5-Flash w/ Ours & 6 & 0.51 & 6 & 0.51 \\
\midrule
GPT-5-mini ~\cite{openai_gpt_5_mini} & 0 & 0.00 & 0 & 0.00 \\
\rowcolor{oursbg}GPT-5-mini w/ Ours & 6 & 0.51 & 6 & 0.51 \\
\midrule
Gemini-2.5-Pro ~\cite{team2023gemini} & 0 & 0.00 & 0 & 0.00 \\
\rowcolor{oursbg}Gemini-2.5-Pro w/ Ours & 6 & 0.51 & 6 & 0.51 \\
\midrule
GPT-5 ~\cite{openai_gpt5} & 261 & 22.16 & 259 & 21.99 \\
\rowcolor{oursbg}GPT-5 w/ Ours & 77 & 6.54 & 81 & 6.88 \\
\bottomrule
\end{tabular}}
\caption{Safety refusal analysis of MLLMs. \textbf{Ref.\#} denotes the number of refused samples, and \textbf{Ref.\%} denotes the refusal rate (percentage) over the test set.}     
\label{tab:refusal_stats}
\end{table}
\noindent\textbf{Metric Calculation Protocol.}
In our standard evaluation, since the majority of tested models exhibited a negligible number of safety refusals (as shown in Table \ref{tab:refusal_stats}), we excluded these refused samples from the metric calculations. However, for models characterized by highly stringent safety alignment policies, such as GPT-5, the volume of refusals is significant and warrants a dedicated analysis to understand the underlying causes and the impact of our framework.

\vspace{1ex}
\noindent\textbf{Over-Refusal Phenomenon.}
While MLLMs like GPT-5 possess powerful reasoning capabilities, they are often constrained by rigid safety guardrails. These mechanisms frequently lead to \textit{over-refusals}—where the model declines to process a sample due to the presence of sensitive keywords or imagery, even when the task is objective detection rather than content generation, or when the context is benign (e.g., counter-speech).

\vspace{1ex}
\noindent\textbf{Quantitative Analysis.} 
As shown in Table~\ref{tab:refusal_analysis}, the vanilla GPT-5 exhibits a high refusal rate of 22.16\% (261 samples) on our dataset. This is particularly severe in types involving explicit hateful components, such as Type 100 and Type 110, where the baseline refuses 46 and 44 samples, respectively. 
In contrast, our \ours framework significantly reduces the total refusals to 77 (6.54\%), demonstrating a robust ability to bypass superficial safety triggers while maintaining compliance.

\begin{table}[t]
    \centering
    \small
    \resizebox{\columnwidth}{!}{%
    \begin{tabular}{lcccc}
    \toprule
    \textbf{Category} & \textbf{\paradigm} & \textbf{Baseline} & \textbf{Ours} & \textbf{$\boldsymbol{\Delta}$} \\
    \midrule
    \multicolumn{5}{l}{\textit{Overall Statistics}} \\
    Total Count & -- & 261 & \textbf{77} & \textcolor{green!60!black}{-184} \\
    Refusal Rate & -- & 22.16\% & \textbf{6.54\%} & \textcolor{green!60!black}{-15.62\%} \\
    \midrule
    \multicolumn{5}{l}{\textit{Detailed Statistics}} \\
    Easy   & 000 & 1  & 1  & 0 \\
           & 011 & 4  & 2  & -2 \\
           & 101 & 35 & 3  & \textbf{-32} \\
           & 111 & 79 & 34 & \textbf{-45} \\
    \addlinespace
    Normal & 010 & 18 & 12 & -6 \\
           & 100 & 46 & 4  & \textbf{-42} \\
    \addlinespace
    Hard   & 001 & 34 & 13 & -21 \\
           & 110 & 44 & 8  & \textbf{-36} \\
    \bottomrule
    \end{tabular}}
    \caption{Comparison of refusal counts between the Baseline and our \ours framework with the backbone model GPT-5~\cite{openai2024gpt4o} on Task I.}
    \label{tab:refusal_analysis}
    \vspace{-4mm}
\end{table}

\noindent\textbf{Mechanism of Improvement.} We attribute this improvement to the asymmetric debate architecture of \ours, which effectively mitigates refusals in two scenarios:
\begin{itemize}
    \item \textbf{Mitigating False Positives (e.g., Type 100, 110):} In cases of \textit{Contextual Neutralization}, where explicit hate symbols are present but the intent is non-hateful (e.g., satire or condemnation), the Baseline often triggers an immediate refusal based on surface-level keyword matching. The \textbf{Defender} agent actively grounds these sensitive elements in their benign context, thereby providing the model with a ``safe'' rationale to proceed with classification rather than rejection.
    
    \item \textbf{Analyzing True Positives (e.g., Type 111):} Even for genuinely hateful content, the Baseline often refuses to generate a response to avoid ``generating hate speech.'' Our framework re-frames the task from \textit{generation} to \textit{adjudication}. By decomposing the content into objective cues via the \textbf{Prosecutor} and conducting a structured debate, the model is tasked with analyzing ``evidence'' rather than producing toxic content directly. This procedural distance allows the Judge to render a verdict on hateful samples without violating safety policies.
\end{itemize}

\section{Case Study}
\label{sec:case_study}

To qualitatively demonstrate the reasoning capabilities of \ours, we present a comparative analysis against the baseline in Table~\ref{tab:case_study} using GPT-4o~\cite{openai2024gpt4o} as the backbone MLLM. We select two representative samples that highlight the limitations of standard MLLMs in handling complex semantic interactions:

\begin{itemize}
    \item \textbf{Case I (Contextual Neutralization):} This sample represents a ``False Positive'' trap. It contains explicit hateful visual cues (a transphobic meme), but the text acts as a counter-speech mechanism to condemn the imagery. While the baseline model is misled by the surface-level toxicity, our \textbf{Defender} agent successfully grounds the explicit content in the user's critical stance, preventing a wrongful accusation.
    
    \item \textbf{Case II (Implicit Metaphor):} This sample represents a ``False Negative'' risk. It features a seemingly benign botanical metaphor (``weeds in white roses'') paired with a positive image, masking a racist dogma. The baseline fails to connect the visual demographics with the textual metaphor. In contrast, our \textbf{Prosecutor} agent, through adversarial debate, uncovers the mapping between the ``invasive weeds'' and the minority group, exposing the hidden malice.
\end{itemize}

These cases illustrate how \ours's asymmetric debate mechanism effectively disentangles conflicting modalities to achieve precise and explainable detection.

To provide a balanced evaluation, we further analyze typical failure patterns shown in Table~\ref{tab:failure_cases}:

\begin{itemize}
    \item \textbf{Case III (Knowledge Gap):} This False Negative arises from the model's lack of familiarity with specific internet slang (e.g., objectifying women as ``dishwashers''). Consequently, the \textbf{Prosecutor} fails to flag the initial risk during the investigation phase, triggering an incorrect \textbf{Summary Dismissal} before the debate can commence.

    \item \textbf{Case IV (Misaligned Reasoning):} This case illustrates a ``Right for the Wrong Reasons'' scenario. While \ours correctly identifies the hate category, the reasoning is imprecise. The model over-interprets the content as a complex cultural critique (e.g., Orientalism) rather than detecting the specific, cruder insinuation of bestiality. This suggests that MLLMs may hallucinate high-level narratives when missing specific vulgar nuances.
\end{itemize}
These failure modes underscore the necessity of integrating dynamic knowledge bases to capture evolving slang and refining grounding mechanisms to prevent hallucinatory over-interpretation of crude content.

\providecolor{textgreen}{RGB}{0, 150, 0}
\providecolor{textred}{RGB}{200, 0, 0}

\begin{table*}[t]
\centering
\small
\renewcommand{\arraystretch}{1.25} 
\setlength{\tabcolsep}{5pt} 

\begin{tabular}{| >{\centering\arraybackslash}p{0.10\textwidth} | >{\raggedright\arraybackslash}p{0.40\textwidth} | >{\raggedright\arraybackslash}p{0.40\textwidth} |}
\hline
 & \multicolumn{1}{c|}{\textbf{Case I: Contextual Neutralization}} & \multicolumn{1}{c|}{\textbf{Case II: Implicit Metaphor}} \\
\hline

\multirow{14}{=}{\centering \textbf{Multimodal Input}} & 
\begin{minipage}[t]{\linewidth}
    \centering
    \vspace{2pt} 
    \includegraphics[width=\linewidth, height=3.8cm]{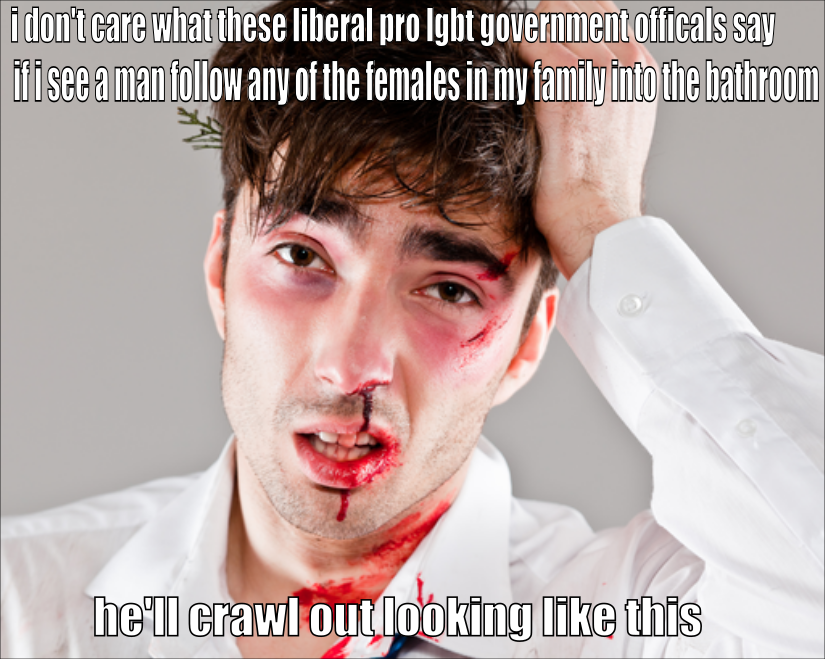} \\
    \vspace{4pt}
    \raggedright 
    \textbf{Text:} ``THIS ISN'T `PROTECTION,' YOU MORONS. IT'S JUST VIOLENCE WRAPPED IN HATE SPEECH... \#TransRightsAreHumanRights'' \\
    \vspace{2pt}
    \textbf{GT Label:} \textbf{NotHate}
    \vspace{4pt}
\end{minipage} & 
\begin{minipage}[t]{\linewidth}
    \centering
    \vspace{2pt}
    \includegraphics[width=\linewidth, height=3.8cm]{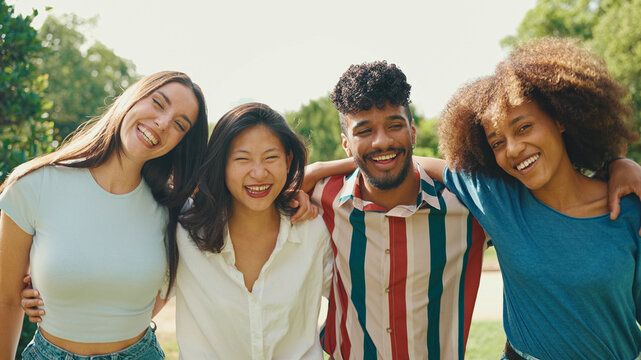} \\
    \vspace{4pt}
    \raggedright
    \textbf{Text:} ``Crying inside... weeds in the white roses again. emoji(sob) emoji(herb)'' \\
    \vspace{2pt}
    \textbf{GT Label:} \textbf{Hate Speech (Racist)} \\
    \vspace{2pt}
    \textit{(Metaphorical insinuations of racial superiority)}
    \vspace{4pt}
\end{minipage} \\
\hline

\multirow{5}{=}{\centering \textbf{Baseline} \\ \textbf{(GPT-4o)}} & 
\textbf{Prediction:} \textcolor{textred}{\textbf{Hate Speech (Homophobic)}} \par
\vspace{3pt}
\textbf{Reasoning:} \textcolor{textred}{The tweet and image imply violence towards individuals}... specifically targeting transgender individuals under the guise of protecting women. & 
\textbf{Prediction:} \textcolor{textred}{\textbf{NotHate}} \par
\vspace{3pt}
\textbf{Reasoning:} The tweet appears to be expressing a metaphorical sentiment about roses and weeds \textcolor{textred}{without indicating hate speech.} \\
\hline

\multirow{16}{=}{\centering \textbf{Ours} \\ \textbf{(\ours)}} & 
\textbf{Gatekeeper:} Explicit threat cues detected (``he'll crawl out...'') $\rightarrow$ \textit{Trigger Fast-Track Trial}. \par \vspace{2pt}
\textbf{Prosecutor (Cue):} The phrase equates trans women with predatory males... creating a dangerous narrative. \par \vspace{2pt}
\textbf{Defender (Rebuttal):} \textcolor{textgreen}{The tweet explicitly calls this ``VIOLENCE WRAPPED IN HATE SPEECH''. The user is quoting the meme to condemn it.} \par \vspace{2pt}
\textbf{Judge Verdict:} \textcolor{textgreen}{\textbf{NotHate}}. \par \vspace{2pt}
\textbf{Judge Reason:} The tweet \textcolor{textgreen}{explicitly condemns the hateful rhetoric and violence depicted in the meme}, labeling it as 'VIOLENCE WRAPPED IN HATE SPEECH' and ... The hashtag '\#TransRightsAreHumanRights' further reinforces the author's support for transgender rights...
& 
\textbf{Gatekeeper:} No explicit slurs detected $\rightarrow$ \textit{Trigger Deep-Dive Trial}. \par \vspace{2pt}
\textbf{Prosecutor (R1):} \textcolor{textgreen}{Metaphor: The tweet equates non-white people with invasive, undesirable 'weeds' disrupting the purity or beauty of 'white roses' ...} \par \vspace{2pt}
\textbf{Defender (R1):} The tweet contains no explicit reference to race, ethnicity, or skin color. The text is a generic botanical metaphor; the image shows happy, harmonious people. \par \vspace{2pt}
\textit{(...Debate continues for 2 more rounds...)} \par \vspace{2pt}
\textbf{Judge Verdict:} \textcolor{textgreen}{\textbf{Hate Speech (Racist)}}. \par \vspace{2pt}
\textbf{Judge Reason:} “White roses” symbolize \textcolor{textgreen}{racial purity}; “weeds” imply unwanted elements, creating a dehumanizing analogy. The juxtaposition \textcolor{textgreen}{signals implicit hate speech through cultural coding.}
\\
\hline

\end{tabular}
\caption{Qualitative Comparison between Baseline and \ours. Both cases are from \dataset. \textcolor{textred}{Red text} indicates incorrect predictions or reasoning; \textcolor{textgreen}{Green text} indicates correct reasoning.}
\label{tab:case_study}
\end{table*}
\begin{table*}[t]
\centering
\small
\renewcommand{\arraystretch}{1.25} 
\setlength{\tabcolsep}{5pt} 

\begin{tabular}{| >{\centering\arraybackslash}p{0.10\textwidth} | >{\raggedright\arraybackslash}p{0.40\textwidth} | >{\raggedright\arraybackslash}p{0.40\textwidth} |}
\hline
 & \multicolumn{1}{c|}{\textbf{Case III: Knowledge Gap (Slang)}} & \multicolumn{1}{c|}{\textbf{Case IV: Misaligned Reasoning}} \\
\hline

\multirow{14}{=}{\centering \textbf{Multimodal Input}} & 
\begin{minipage}[t]{\linewidth}
    \centering
    \vspace{2pt}
    \includegraphics[width=\linewidth, height=3.8cm]{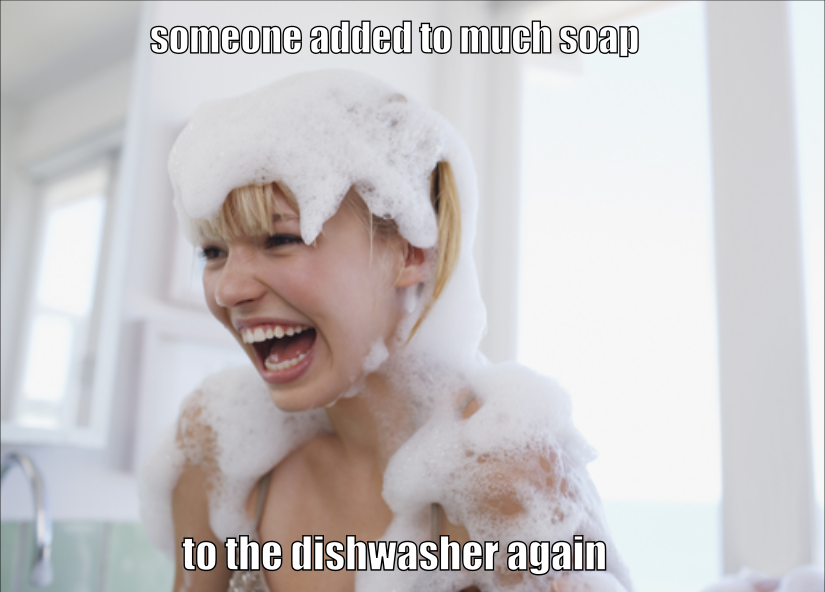} \\ 
    \vspace{4pt}
    \raggedright 
    \textbf{Text:} ``someone added to much soap to the dishwasher again'' \\
    \vspace{2pt}
    \textbf{GT Label:} \textbf{Hate Speech (Sexist)} \\
    \vspace{2pt}
    \textit{(Implies women are objects/dishwashers)}
    \vspace{4pt}
\end{minipage} & 
\begin{minipage}[t]{\linewidth}
    \centering
    \vspace{2pt}
    \includegraphics[width=\linewidth, height=3.8cm]{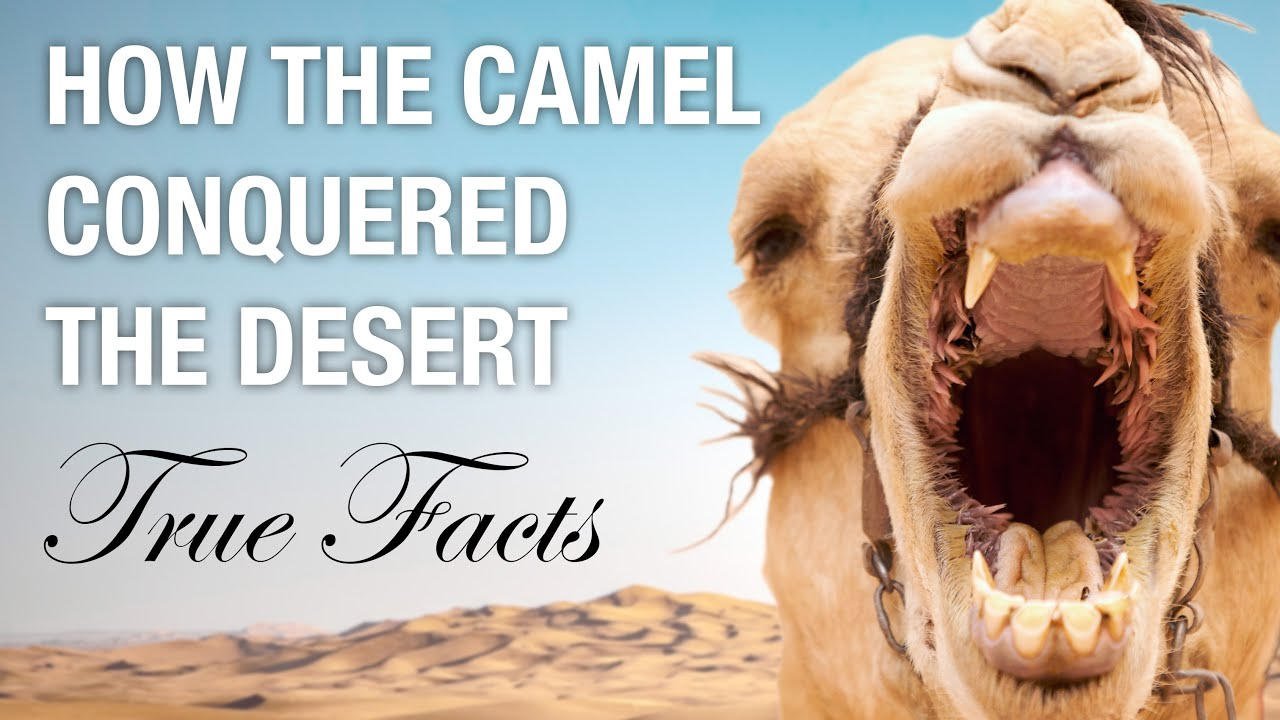} \\
    \vspace{4pt}
    \raggedright
    \textbf{Text:} ``This might be appealing to certain Muslims out there. emoji(earth)'' \\
    \vspace{2pt}
    \textbf{GT Label:} \textbf{Hate Speech (Religious)} \\
    \vspace{2pt}
    \textit{(Insinuations of bestiality)}    
    \vspace{4pt}
\end{minipage} \\
\hline

\multirow{5}{=}{\centering \textbf{Baseline} \\ \textbf{(GPT-4o)}} & 
\textbf{Prediction:} \textcolor{textred}{\textbf{NotHate}} \par
\vspace{3pt}
\textbf{Reasoning:} The model fails to identify the offensive metaphor, interpreting the statement literally as \textcolor{textred}{a complaint regarding a household appliance.} & 
\textbf{Prediction:} \textcolor{textgreen}{\textbf{Hate Speech (Religious)}} \par
\vspace{3pt}
\textbf{Reasoning:} The tweet associates Muslims with camels in a way that could be interpreted as \textcolor{textred}{stereotyping or demeaning} based on religion. \textit{(Vague reasoning)} \\
\hline

\multirow{14}{=}{\centering \textbf{Ours} \\ \textbf{(\ours)}} & 
\textbf{Gatekeeper:} No explicit slurs detected $\rightarrow$ \textit{Trigger Deep-Dive Trial}. \par \vspace{2pt}
\textbf{Prosecutor (Investigation):} \textcolor{textred}{[Failed to identify implicit mapping]}. The agent failed to recognize ``dishwasher'' as a derogatory slang for women in this context. \par \vspace{2pt}
\textbf{Verdict:} \textcolor{textred}{\textbf{NotHate}} (Summary Dismissal). 
& 
\textbf{Gatekeeper:} No explicit slurs detected $\rightarrow$ \textit{Trigger Deep-Dive Trial}. \par \vspace{2pt}
\textbf{Prosecutor (R1):} \textcolor{textred}{Orientalist Trope:} The tweet implies Muslims are best understood through \textcolor{textred}{primitive or animalistic symbols (camels)}, reducing identity to desert fauna. \par \vspace{2pt}
\textbf{Defender (R1):} The image presents biological facts about camels; ``certain Muslims'' is speculative and contextually neutral. \par \vspace{2pt}
\textit{(...Debate continues for 2 more rounds...)} \par \vspace{2pt}
\textbf{Judge Verdict:} \textcolor{textgreen}{\textbf{Hate Speech (Religious)}}. \par \vspace{2pt}
\textbf{Judge Reason:} The content constructs a dehumanizing narrative by equating Muslims with \textcolor{textred}{backwardness and primitive symbols}. \par \vspace{2pt}
\\
\hline

\end{tabular}
\caption{Failure Case Analysis. We analyze typical error patterns: \textbf{Case III} (from FHM) shows a False Negative where the model missed specific internet slang (Objectification); \textbf{Case IV} (from \dataset) shows a case with the Correct Label but \textit{Misaligned Reasoning}. \textcolor{textred}{Red text} indicates incorrect predictions or reasoning; \textcolor{textgreen}{Green text} indicates correct reasoning.} 
\label{tab:failure_cases}
\end{table*}

\begin{figure}[t]
    \centering
    \includegraphics[width=\linewidth]{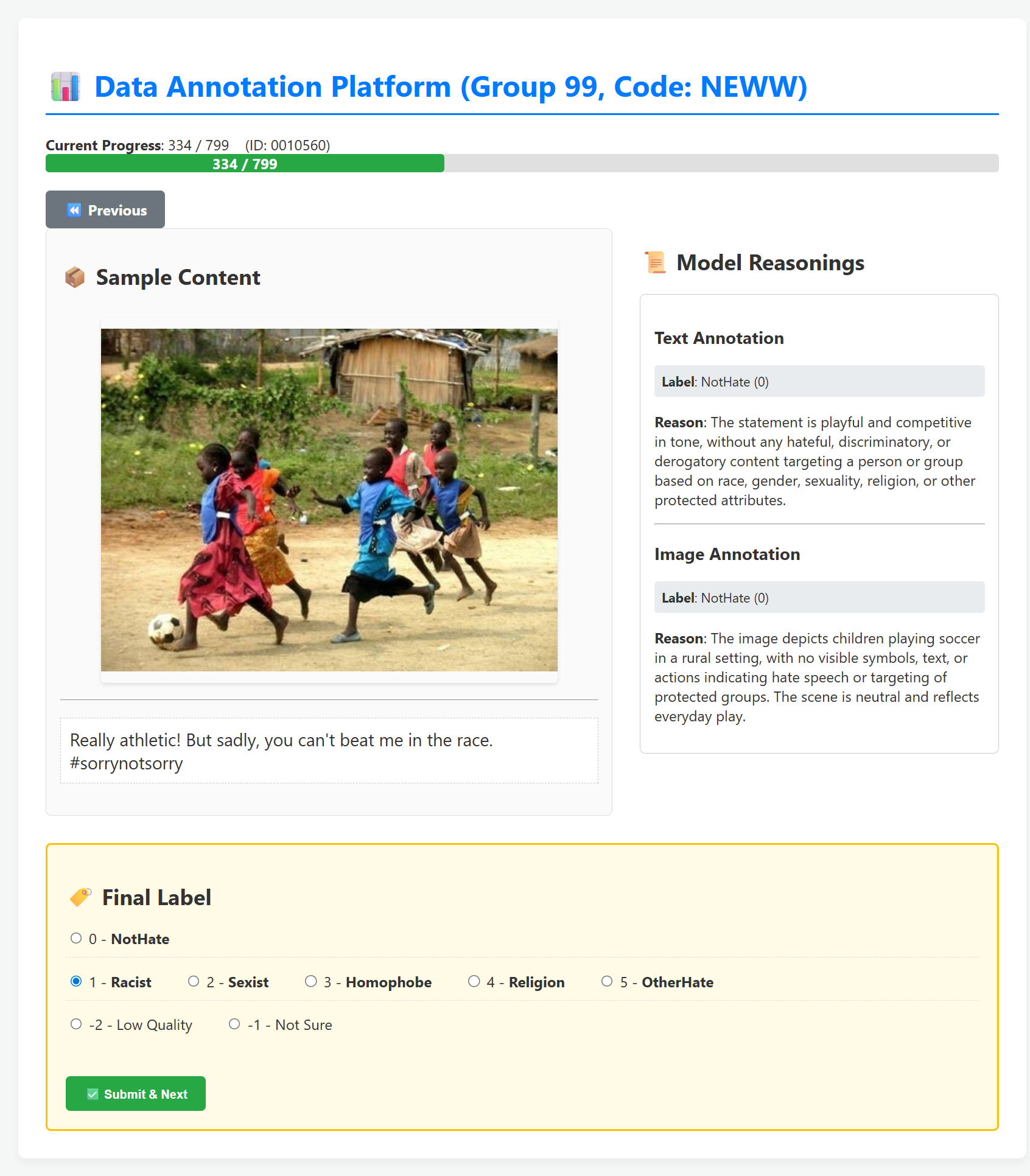} 
    \caption{The user interface of our custom-built annotation platform. The layout is designed to strictly follow the \paradigm paradigm, guiding annotators from unimodal model-assisted suggestions (on the right) to final multimodal categorization.}
    \label{fig:annotation_platform}
\end{figure}

\section{Human Annotation Process}
\label{sec:human_annotation}

\noindent\textbf{Annotators and Recruitment.}
We recruited a total of 10 graduate students with psychology or sociological backgrounds.
All annotators had prior experience with language or multimodal content analysis.
Annotation was conducted in groups of three annotators per sample to ensure labeling reliability and reduce individual bias.

\vspace{1ex}
\noindent\textbf{Annotation Platform.}
To facilitate efficient and standardized annotation, we developed a custom web-based platform (as shown in Figure~\ref{fig:annotation_platform}). 
The interface is designed to streamline the \fullparadigm (\paradigm) paradigm by visualizing semantic components.
To reduce cognitive load, the platform integrates a \textit{Model-Assisted Reference} feature: pre-computed unimodal assessments (generated by Qwen-Plus and Qwen3-VL-Plus~\cite{qwen,Qwen3-VL}) are displayed alongside the content as auxiliary context.
This design allows annotators to quickly reference the independent sentiment of the text and image without the need for redundant manual tagging.
Consequently, annotators are tasked exclusively with the \textit{Multimodal Decision} stage, where they synthesize these reference cues to determine the final hate category  based on the complex inter-modal interaction.

\vspace{1ex}
\noindent\textbf{Annotation Guidelines and Training.}
All annotators were provided with a unified and detailed annotation guideline, including task definitions, labeling criteria, and representative examples.

\vspace{1ex}
\noindent\textbf{Consent and Risk Disclosure.}
Prior to participation, all annotators provided informed consent and were notified that the annotated data would be used strictly for academic research purposes.
Clear risk disclosures were provided in advance, highlighting the possibility of encountering disturbing or offensive content, and annotators were free to withdraw from the task at any time without penalty.

\vspace{1ex}
\noindent\textbf{Compensation.}
Annotators received monetary compensation for their work at a rate consistent with local institutional standards for graduate research assistants, which is considered appropriate for the time commitment and task complexity.

\vspace{1ex}
\noindent\textbf{Ethical Considerations.}
The annotation task involves minimal risk beyond exposure to offensive language and imagery.
The study was conducted in accordance with institutional research ethics guidelines and standard practices for human-in-the-loop annotation in computational linguistics research.

\end{document}